\documentclass[letterpaper, 10 pt, conference]{ieeeconf}  % Comment this line out if you need a4paper

\IEEEoverridecommandlockouts                              % This command is only needed if 
\usepackage[top=60pt, left=48pt, right=48pt, bottom=43pt]{geometry}
\usepackage[utf8]{inputenc}
\usepackage{xcolor}
\usepackage{graphicx}
\usepackage{subcaption}
\usepackage{lipsum}  
\usepackage{balance}
\usepackage{rotating}
\usepackage{transparent}
\usepackage[font=small]{caption}
\usepackage{arydshln}
\usepackage{epstopdf}
\usepackage{array}
\usepackage{tabularx}
\usepackage{makecell}
\usepackage{multirow}
\usepackage{setspace}
\usepackage{booktabs}
\usepackage{amsfonts}
\makeatletter
\let\NAT@parse\undefined
\makeatother          
\usepackage{standalone}

\usepackage{tikz}
\usetikzlibrary{patterns,positioning,arrows,arrows.meta,calc,shapes,pgfplots.groupplots,fit,backgrounds}
\title{\LARGE \bf
Towards a Robust Sensor Fusion Step for 3D Object \\ Detection on Corrupted Data
}

\author{Viktor Kårefjärds*$^{1}$, Maciej K. Wozniak*$^{1}$, Marko Thiel$^{2}$, Patric Jensfelt$^{1}$% <-this % stops a space
\thanks{*Both authors contributed equally to this work}% <-this % stops a space
\thanks{$^{1}$Viktor Kårefjärds, Maciej K. Wozniak, and Patric Jensfelt are with the Division of Robotics, Perception, and Learning, KTH Royal Institute of Technology, Stockholm, Sweden}
\thanks{$^{2}$Marko Thiel is with TU Hamburg, Germany}
}

\begin{document}

\maketitle
\thispagestyle{empty}
\pagestyle{empty}

\begin{abstract}
Multimodal sensor fusion methods for 3D object detection have been revolutionizing the autonomous driving research field. Nevertheless, most of these methods heavily rely on dense LiDAR data and accurately calibrated sensors which is often not the case in real-world scenarios. Data from LiDAR and cameras often come misaligned due to the miscalibration, decalibration, or different frequencies of the sensors. Additionally, some parts of the LiDAR data may be occluded and parts of the data may be missing due to hardware malfunction or weather conditions. This work presents a novel \textit{fusion step} that addresses data corruptions and makes sensor fusion for 3D object detection more robust. Through extensive experiments, we demonstrate that our 
method performs on par with state-of-the-art approaches on normal data and outperforms them on misaligned data.
%method surpasses most state-of-the-art approaches, achieving superior results on both corrupted and uncorrupted data.
\end{abstract}

\maketitle
\thispagestyle{empty}
\pagestyle{empty}

\section{Introduction}
\label{sec:intro}

% Recently, advances in computer vision through the application of neural networks, enabled by increased computing power, have made self-driving vehicles highly relevant. 

Self-driving cars must understand their own surroundings, such as vehicles, pedestrians, or cyclists, as well as their pose to further estimate the velocity or future trajectory of moving objects and plan their own movement accordingly. 3D object detection is often used to retain this semantic information about the environment~\cite{khoche2022semantic}.

3D object detection methods rely on different types of data collected using LiDAR~\cite{PointPillars} or RGB cameras~\cite{FCOS3D}, or a combination of those~\cite{Bevfusion-Liang}. 

% RGB-only methods often depend on depth estimation and are usually outperformed by LiDAR and fusion methods when it comes to performance. 

% The methods embracing the use of a LiDAR sensor operate on the 3D clouds of reflection points captured by the sensor called \textit{point clouds}. 

% Finally, fusion methods combine both sensor data to improve the performance and are currently obtaining the highest results.

% The extra dimension (depth) in LiDAR and fusion methods makes convolutional neural networks (CNN) difficult to apply directly due to the sliding-window kernel approach, convolution becomes computationally expensive since it must operate in three-dimensional space~\cite{SECOND}. A paradigm to avoid 3D convolutions is the use of various point cloud projections and the Bird's Eye View (BEV) has been lately the most prominent solution~\cite{BirdNet}. The idea is to view the point cloud from above, and thus perform a dimension reduction on the inherently sparse LiDAR data. This is motivated by the fact that detectable objects (e.g. cars) in the outdoor environment do not tend to be stacked on top of one another.

Although these methods are capable of achieving impressive results, they often heavily rely on dense LiDAR data and accurately calibrated sensors. Unfortunately, this is often not the case in the real-world scenario and there are different ways the input data can be corrupted. Data from LiDAR and camera often comes misaligned due to poor initial miscalibration or decalibration throughout the vehicle movement~\cite{das2023observability} as well as different frequencies or latencies of the sensors~\cite{transfusion}. Additionally, some areas of the LiDAR may be occluded and parts of the data may be missing due to hardware malfunction, weather conditions, or reflective surfaces~\cite{nguyen2022ntu}. Furthermore, robotic platforms use LiDAR sensors with different resolutions and even though some methods claim to achieve good results on any given data set, they often underperform when tested on domains they not are trained on (e.g. trained on 64-layer LiDAR, tested on 16 layers)~\cite{wozniak2023applying}. %,yang2021st3d}. 

% The cases described above case decrease in performance of 3D object detection models, thus, we refer to these types of data as \textit{corrupted}. 

These different cases of data corruption lead to a considerable decline in performance for state-of-the-art single-modality 3D object detection methods, making them unreliable in real-world scenarios. While multimodal fusion methods are also impacted by these issues, they are more resilient than single-modality approaches. Nevertheless, their effectiveness and robustness heavily depend on where and how the information is fused within the model. 

% combine the data from RGB images and point clouds, to achieve better accuracy in 3D object detection than single modality approaches, they are also affected by data corruption. 

For example, early fusion, which combines modalities almost at the input level, is more prone to corrupted data. On the other hand, deep fusion can be more robust as it allows the network to learn more abstract representations from multiple modalities, potentially mitigating the impact of corrupted or missing information.

Similarly, the way in which the information are fused, what we refer to as \textit{fusion step}, can exhibit different levels of sensitivity to corrupted data. For instance, combining features from different modalities directly (e.g. simply concatenating them together) makes the model highly susceptible to corruption in any of the modalities, whereas using convolution operation to fuse the data can improve handling noise and misalignment.

This work aims to explore how multi-modal fusion could be performed to ensure robustness to corrupted data, required for a practical application, since in the real world we rarely operate on dense data without missing information and with perfectly calibrated sensors. Our main contribution is a \textit{novel fusion step} for 3D object detection that outperforms other proposed state-of-the-art fusion methods on data from miscalibrated sensors and achieves similar or better results when it comes to LiDAR layer removal and point cloud reduction. We also provide the code for our benchmarking experiments, so that others can reproduce our results as well as test their methods on corrupted sensor data, making multimodal fusion more reliable in real-world scenarios \textit{https://github.com/ViktorKare/bevf}. 

\section{Sensor Fusion for 3D object detection}
\label{sec:related}

% \begin{figure}[!t]
%      \centering
%      % \begin{adjustbox}{center}
%      \begin{subfigure}[b]{0.6\columnwidth}
%          \centering
%          \includegraphics[width=\textwidth]{figures/typical_fusion_network.pdf}
%          \caption{Example of sensor fusion network architecture for 3D object detection.}
%          \label{fig:fus_ex_flow}
%      \end{subfigure}
%      \hfill
%      \begin{subfigure}[b]{0.35\columnwidth}
%          \centering
%          \includegraphics[width=\textwidth]{figures/fusion_taxonomy.pdf}
%          \caption{Example of sensor fusion taxonomy.}
%          \label{fig:fus_tax}
%      \end{subfigure}
%      % \end{adjustbox}
%      \caption{Fusion-based architecture and terminology.}
% \end{figure}

One of the main benefits of the RGB-camera in an object detection scene is the semantic-rich nature of the data. Each image holds hundreds of thousands that are closely semantically related. LiDAR-data on the other hand does not carry semantic information and even expensive high-resolution LiDAR sensors capture point clouds that are much sparser in comparison to RGB-cameras, but their advantage is the ability to directly retain geometric information about the scene. This section describes \textit{multimodal fusion} approaches used in 3D object detection that integrate camera and LiDAR to leverage the strengths of both sensors.
\begin{figure*}[!t]
     \centering
     \begin{subfigure}[b]{0.32\textwidth}
         \centering
        \includegraphics[width=\textwidth]{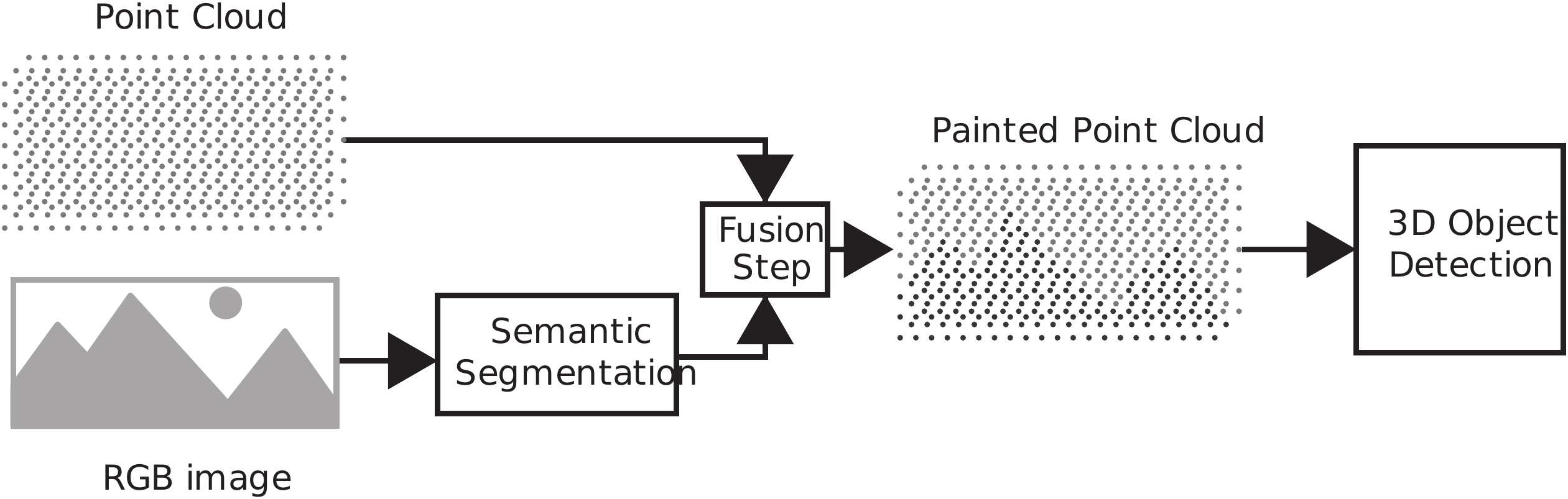}
        \caption{Early fusion architecture.}
        \label{fig:early_fusion}
     \end{subfigure}
     \hfill
     \begin{subfigure}[b]{0.32\textwidth}
        \centering
        \includegraphics[width=\textwidth]{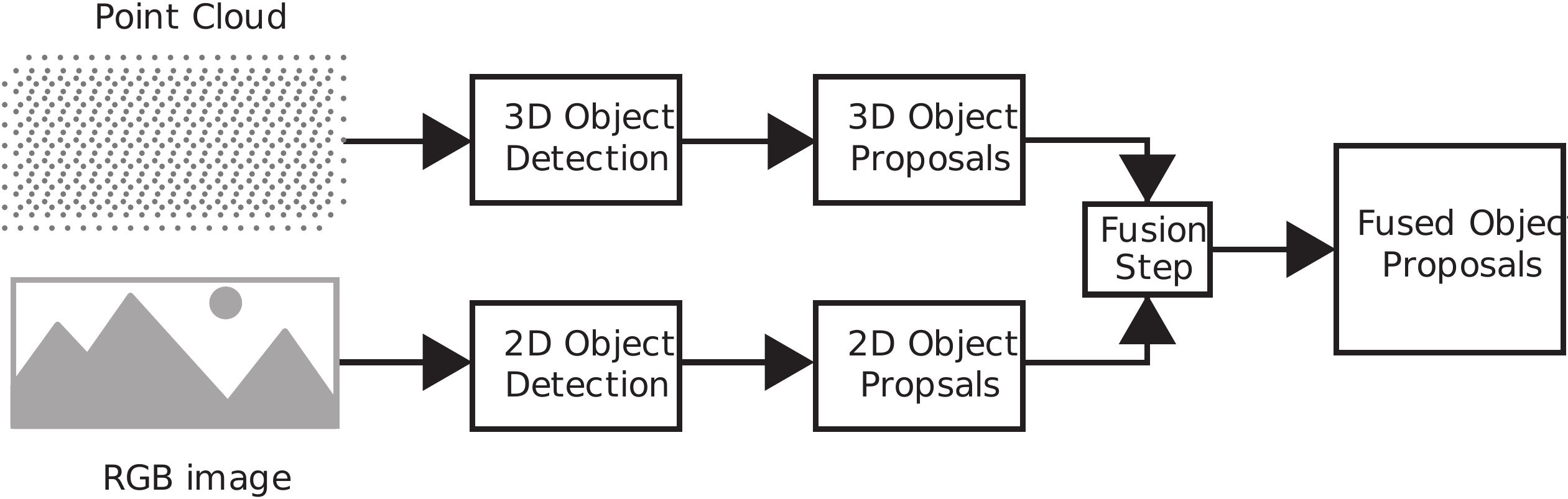}
        \caption{Late fusion architecture.}
        \label{fig:late_fusion}
     \end{subfigure}
     \hfill
     \begin{subfigure}[b]{0.32\textwidth}
        \centering
        \includegraphics[width=\textwidth]{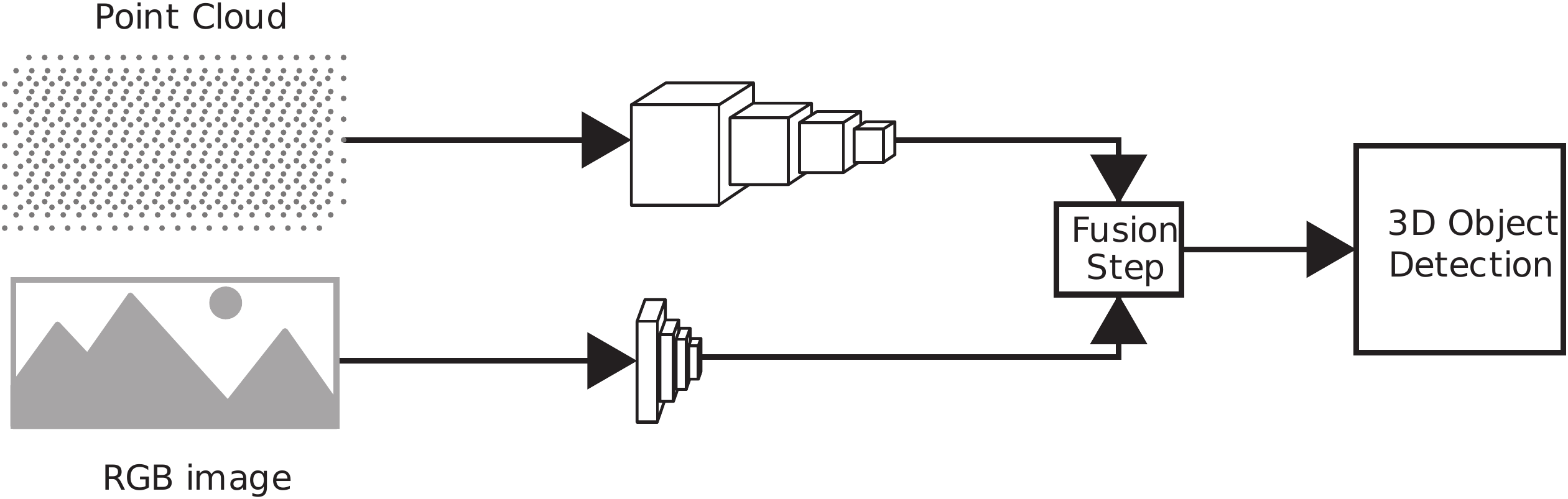}
        \caption{Deep fusion architecture.}
        \label{fig:deep_fusion}
     \end{subfigure}
     \caption{Deep neural network fusion architectures.}
     \label{fig:fusionarch}
\end{figure*}

\begin{figure*}[t!]
     \centering
     \begin{subfigure}[b]{0.32\textwidth}
        \centering
        \includegraphics[width=0.75\textwidth]{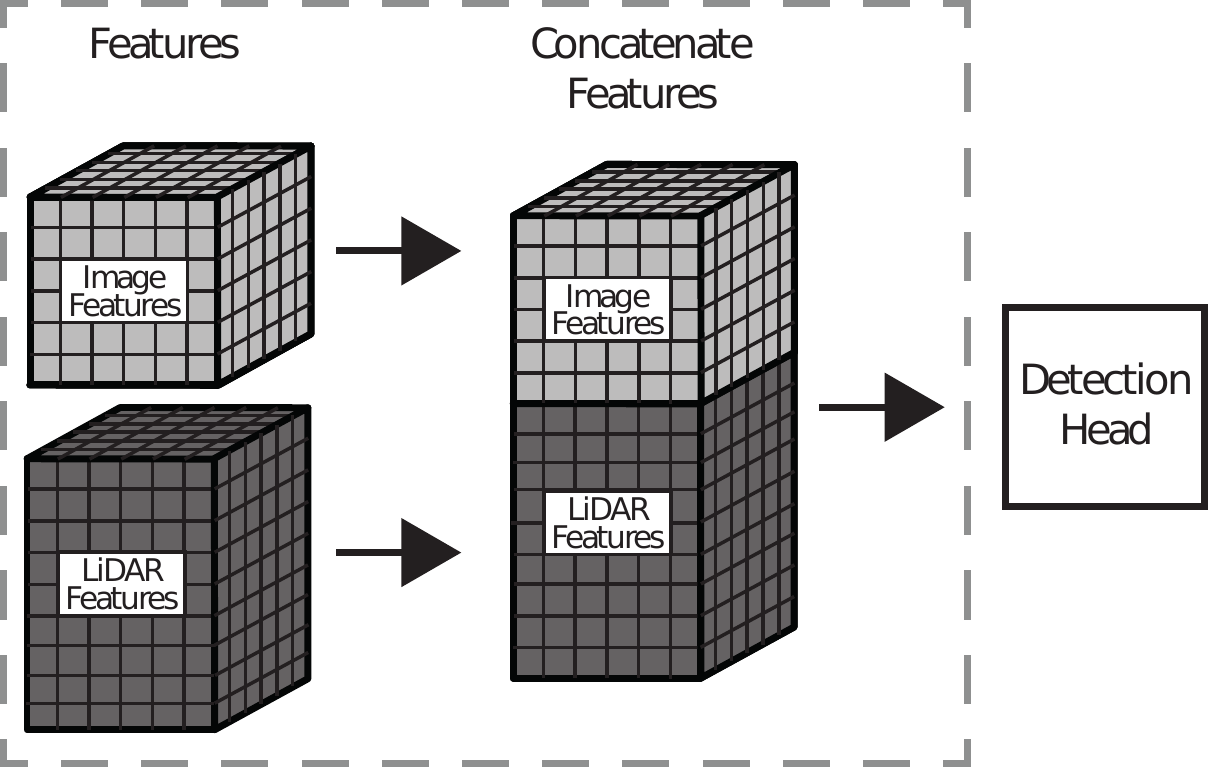}
        \caption{\textit{Concatenation} fusion step.}
        \label{fig:fusion_step_concat}
     \end{subfigure}
     \hfill
     \begin{subfigure}[b]{0.32\textwidth}
        \centering
        \includegraphics[width=\textwidth]{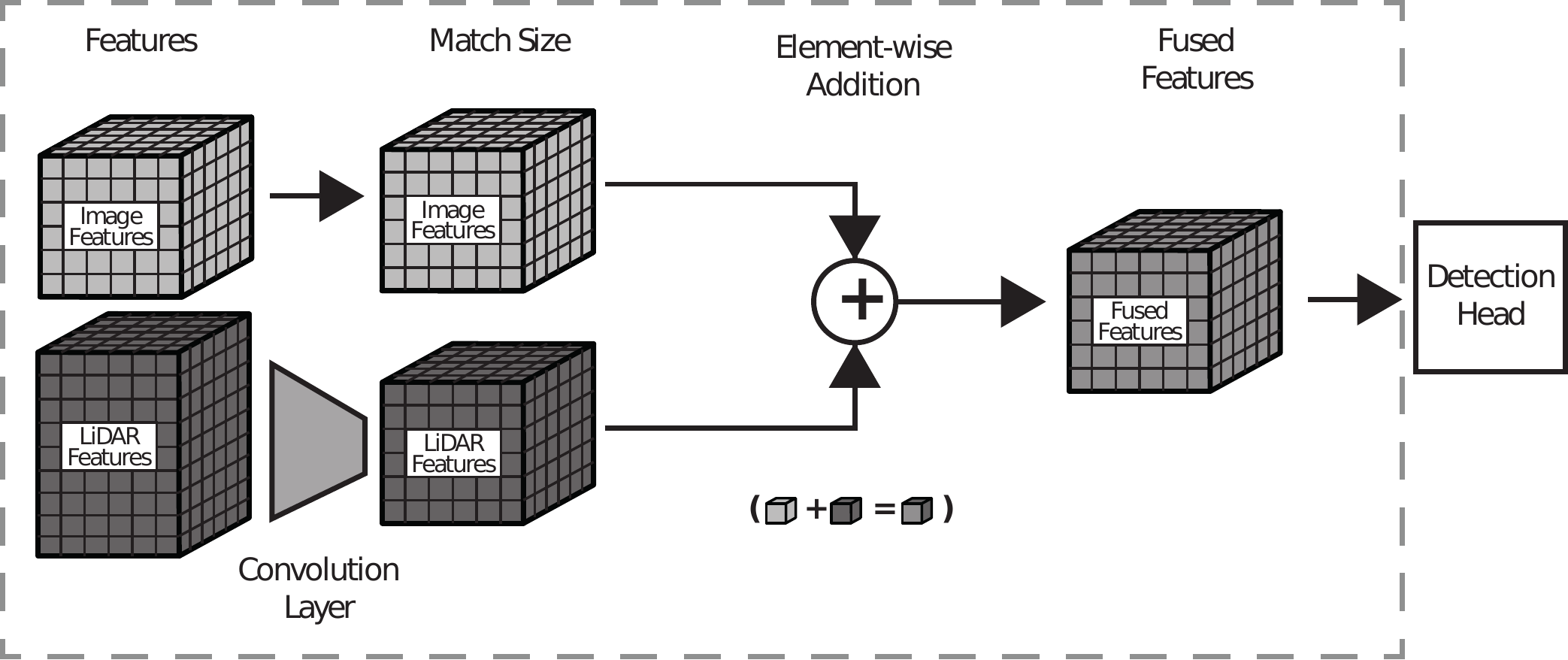}
        \caption{\textit{Element-wise addition} fusion step.}
        \label{fig:fusion_step_men-wise_add}
     \end{subfigure}
     \hfill
     \begin{subfigure}[b]{0.32\textwidth}
        \centering
        \includegraphics[width=\textwidth]{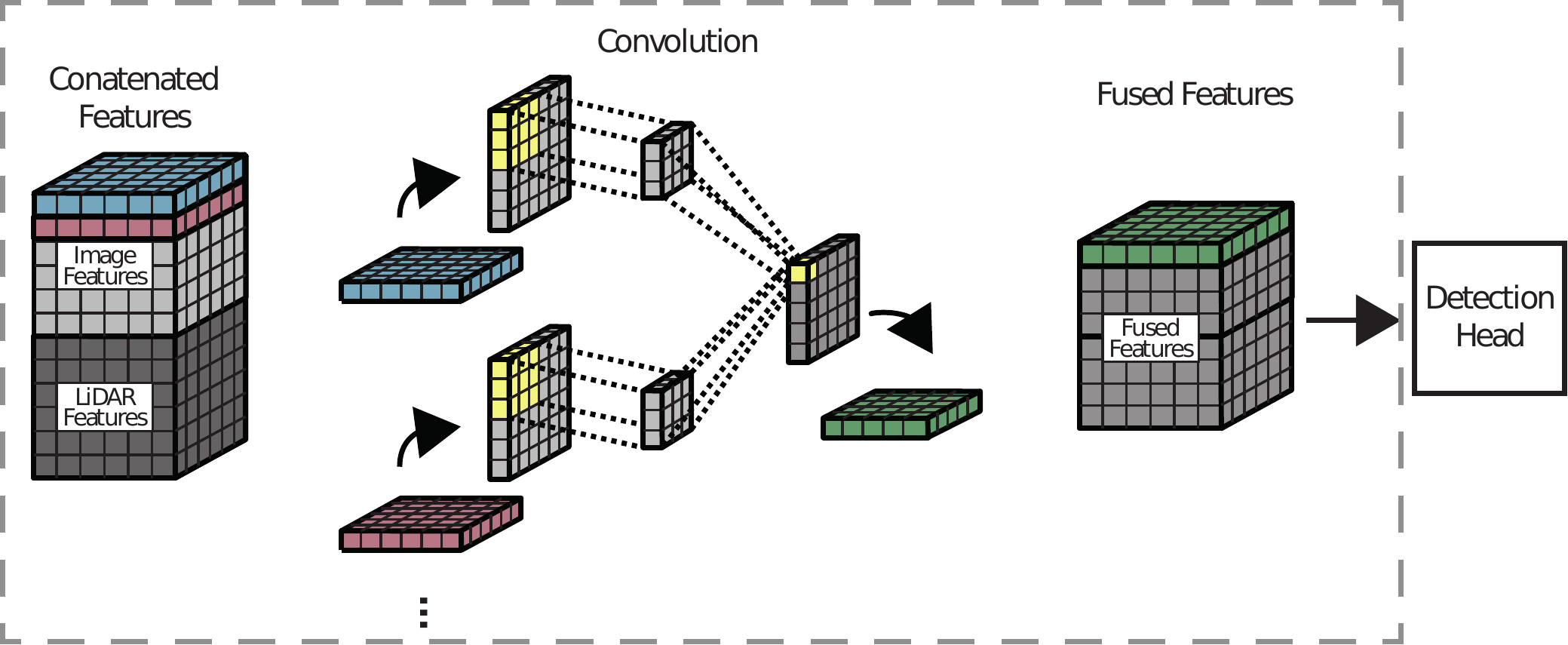}
        \caption{\textit{Convolution} fusion step.}
        \label{fig:fusion_step_conv}
     \end{subfigure}
     \caption{Basic fusion steps. These methods are often used as an intermediate \textit{stage} in other, more advanced fusion steps.}
     \label{fig:stepbasic}
\end{figure*}

% Other papers describing novel 3D object detection multimodal fusion methods often fail to present the methods and their architecture in depth so that the reader can easily understand where and how the sensor fusion happens in the model. 

We describe common fusion architectures (where the fusion happens in the network) as well as the fusion steps (how the information is combined together) to support our analysis and design.

% We do this so that any researcher who has an understanding of the 3D object detection field, will be able to get a basic understanding of the merits and demerits of multimodal fusion approaches.

\subsection{Multi-modal Fusion Architectures}
\label{subsec:fus_arch}

Multi-modal 3D object detection models fuse data at various stages of the network \cite{MV3D,MVXNET,BEVFusion,Bevfusion-Liang,clocs}. The type of fusion can be categorized in accordance with the level at which the fusion is performed~\cite{huang2022multi}. 

% A work by Bai et al. \cite{transfusion} describes three categories point-level, proposal-level, and, results-level. Another paper \cite{li2022deepfusion}, refers to point decoration and mid-level fusion. Finally, \cite{pfeuffer2018optimal} simply uses early-fusion, middle-fusion, and, late-fusion. 

\textbf{\textit{Early-fusion}}, shown in Figure \ref{fig:early_fusion}, represents a data-level fusion, where the modalities are combined before any significant feature encoding. Strategies include a raw or processed image-to-point projection. Methods like \cite{vora2020pointpainting,MVXNET} fuse semantic segmentation of RGB image with the LiDAR point cloud. The main difficulty with this approach is the significant difference between the two data modalities at the early stage in the pipeline, which can make these methods prone to noise and corruption. 
% projecting the point cloud to a 2D plane or a

\textbf{\textit{Late-fusion}}, shown in Figure \ref{fig:late_fusion}, operates at the object level. In this type of approach, the image and LiDAR pipelines are largely isolated until proposals (e.g. bounding boxes) from each branch are generated~\cite{clocs}. Fusion here focuses on integrating proposals from two branches and incorporating features, such as confidence scores, to generate the model the final IoU score. 

\textbf{\textit{Deep-fusion}}, shown in Figure \ref{fig:deep_fusion}, performs fusion at the feature level where both modalities are first encoded by a neural network backbone (e.g. SwinTransformer for camera \cite{SwinTransformer} and PointNet for LiDAR \cite{PointNet}) into the feature space and then combines together on the feature level. As Liu et al., Liang et al., or Bai et al. showed~\cite{BEVFusion,Bevfusion-Liang,transfusion}, this approach is the most robust towards different disturbances in the data and performs significantly better than early or late fusion, however, comes with an inference speed penalty.
% \begin{figure}[!h]
%     \centering
%     \includegraphics[width=0.8\columnwidth]{figures/late.pdf}
%     \caption{Late fusion architecture.}
%     \label{fig:late_fusion}
% \end{figure}

% Another type is \textbf{\textit{Asymmetry-fusion}}. This category includes the methods that fuse information from the two branches at different levels. Especially, in this work, we defined this type of fusion as approaches that fuse object-level from one branch, at the data or feature level from other branches~\cite{transfusion}. Note how this can be applied in both ways for the two modalities LiDAR and RGB camera image. 

% The final category, \textbf{\textit{weak fusion}}, does not aim to fuse information over the modalities. It uses the information from one branch to improve the results from another, but the information transfer is limited. 

% Frustum PointNets~\cite{FrustumPointNets} is a work of this type where fusion focuses on the regions of interest proposals. This method uses 2D object detection methods to generate object region proposals, the proposals are then extruded to a 3D-frustum, the cone of vision captured by a camera. The LiDAR point in this frustum is then used to predict a full 3D bounding box for the object. This type of approach can be applied with good results, however, no actual fusion of the two modalities occurs, thus we do not consider these methods in our experiments.

\subsection{Fusion step}
% In \ref{subsec:fus_arch} we presented different architectures and stages when  fusion can happen in the network. However, T
There are many ways how the information can be fused together. We refer to this operation as a \textit{fusion step} (marked in Figure \ref{fig:fusionarch}) and discuss it in detail in this section.
% for deep-fusion architectures.   

The simplest way to combine camera and point cloud features tensors is through \textbf{\textit{concatenation}} resulting in a large feature tensor, shown in Figure \ref{fig:fusion_step_concat}. This leaves the dense detection head with the unfused data, and consequently, the head learns to use the two modalities to perform detection. The potential strength of this fusion step approach is the low information loss. With a sophisticated detection head, the choice to keep the two feature spaces separated could be an advantage. This approach is used as an intermediate operation in most of the methods, however, PointPainting~\cite{pointpainting} uses it as a main part of the fusion-step block.

% \begin{figure}[!h]
%     \centering
%     \includegraphics[width=0.77\columnwidth]{figures/fusion_step/concatenate_step.pdf}
%     \caption{\textit{Concatenation} fusion step.}
%     \label{fig:fusion_step_concat}
% \end{figure}

The \textbf{\textit{element-wise addition}} fusion step is feature-to-feature addition, where each feature value in the LiDAR tensor is added with the respective value in the image feature tensor to create fused features of the same size, see Figure \ref{fig:fusion_step_men-wise_add}. Note how the feature tensors from the two data streams must be the same size to perform the step. This fusion step can be found in e.g. in MVXNet~\cite{MVXNET}. Although no point-wise operations are performed in this fusion step as a consequence of the voxelized feature space.
% , weselect to use the ReLU function. 

% \begin{figure}[!h]
%     \centering
%     \includegraphics[width=0.8\columnwidth]{figures/fusion_step/mean-wise_add.pdf}
%     \caption{\textit{Element-wise addition} fusion step.}
%     \label{fig:fusion_step_men-wise_add}
% \end{figure}

The next methods use different types of neural network layers to fuse the signals. The \textbf{\textit{convolution}} fusion step operates on a concatenated tensor originating from the two separated data sources. Once concatenated, the fusion step is not different from a standard channel-reducing convolution. A kernel (sliding window) operates on the feature space by sliding over the larger concatenated feature tensor (see Figure \ref{fig:fusion_step_conv}). The operation is then repeated to include all combinations. Notice how the number of kernels is selected in such a way that the resulting fused feature tensors are reduced along the channel dimension. The \textit{convolution} is used in many methods as an intermediate or standalone fusion step in many SOTA methods~\cite{BEVFusion,Bevfusion-Liang}.

% Next, let's consider more advanced fusion steps that build on the previously introduced approaches.

We can think about previous fusion steps, shown in Figure \ref{fig:stepbasic}, as basic building blocks. The following, more advanced approaches, often use them as intermediate operations. 
\begin{figure}[!t]
    \centering
    \includegraphics[width=0.85\columnwidth]{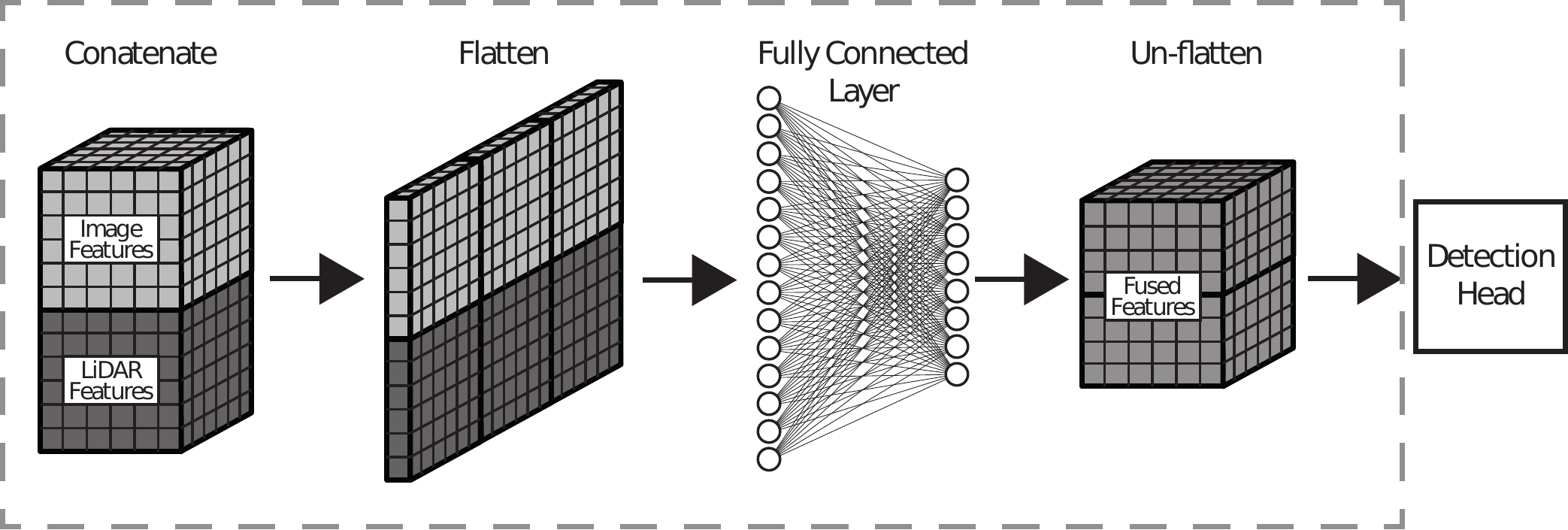}
    \caption{\textit{Fully connected} fusion step.}
    \label{fig:step_fc}
\end{figure}

\textbf{\textit{Fully connected}} (showed in Figure \ref{fig:step_fc}) starts from concatenating two  feature tensors from the image and point cloud into one spatial dimension. Next, they are flattened along the other spatial dimension and used as input for the fully connected layer. This fusion module includes batch normalization and an activation function after the fully connected layer, before feeding this information into the detection head. PointFusion uses a similar approach with multiple fully connected layers (i.e. multi-layer perceptron -- MLP)~\cite{pointfusion}.

\begin{figure}[!b]
  \begin{center}
    \includegraphics[width=0.85\columnwidth]{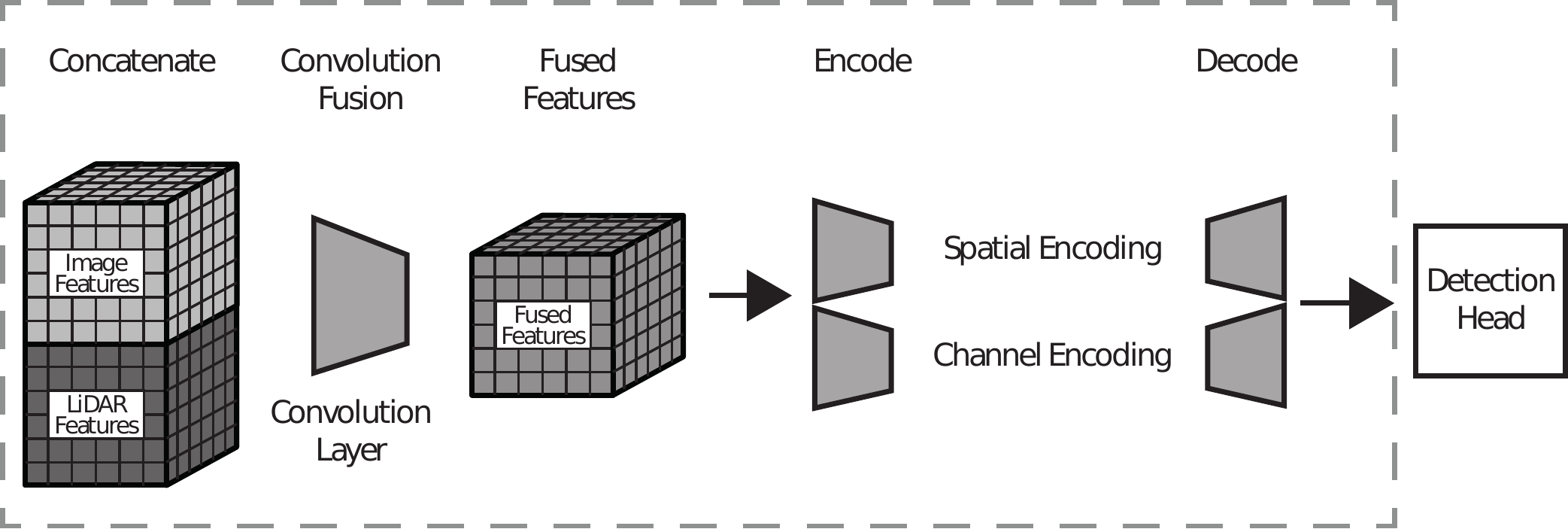}
  \end{center}
  \caption{\textit{Encoder and decoder} fusion step.}
  \label{fig:fusion_step_conv_encodedecode}
\end{figure}

% \begin{figure}[!h]
%   \begin{center}
%     \includegraphics[width=0.8\columnwidth]{figures/fusion_step/convolution.pdf}
%   \end{center}
%   \caption{\textit{Convolution} fusion step.}
%   \label{fig:fusion_step_conv}
% \end{figure}

Zhijian et al.~\cite{BEVFusion} proposed the \textbf{\textit{encoder-decoder}} fusion step to address spatial and channel misalignment. We refer to \textit{channel misalignment} as a misalignment between features in the LiDAR and camera feature spaces in the channel direction. As shown in Figure \ref{fig:fusion_step_conv_encodedecode} a small encoder-decoder network module was added after the convolution fusion step. 

First, channel-wise encoding is used where the channel space is encoded to a smaller space, and then this is decoded to upscale it to the original channel size. The idea behind this step is to target any channel-wise misalignment. 

The second parallel step is spatial encoding and decoding where the spatial dimensions are encoded to a smaller space and in a similar way, a decoder is applied in succession to restore the original dimensions. This feature aligning encoding-decoding is applied to account for misalignment but it also comes with more learnable parameters. 

Additionally, two-way encoding does not have any non-encoded information pass through (skip connection) that helps perceive non-encoded information and does not share the information between the channels.

\begin{figure}[!b]
  \begin{center}
    \includegraphics[width=0.85\columnwidth]{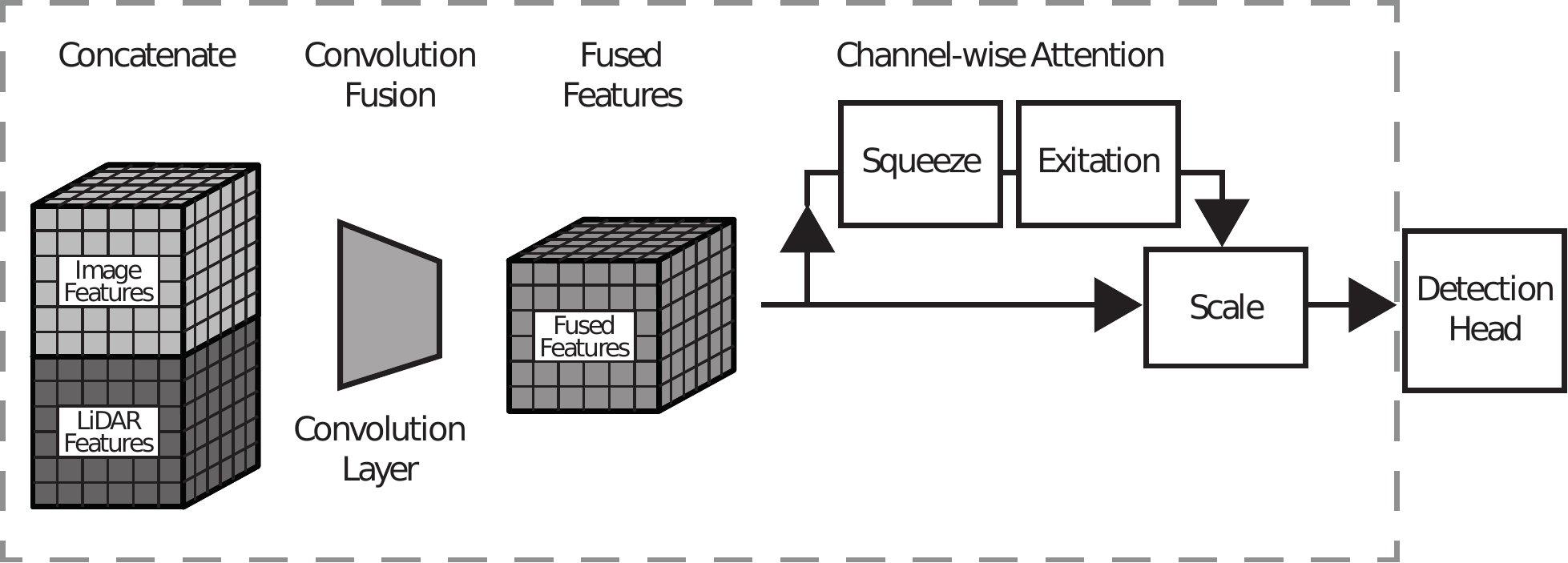}
  \end{center}
  \caption{\textit{SE block} fusion step.}
  \label{fig:se_fusion}
\end{figure}

Hu et al. proposed the \textit{\textbf{Squeeze-and-Excitation}} (SE) fusion step~\cite{squeeze-and-excitation}, showed in Figure \ref{fig:se_fusion}.

The \textit{squeeze} step makes a global average pooling to aggregate features in order to create channel-wise descriptors. The \textit{excitation} step then uses fully connected layers to produce channel-wise activations that are applied to the map of features. The output of the SE block can then be used to scale the convolutional features, according to information value. Thus, essential interdependencies can be enhanced. 
% SE block is used in SOTA fusion approaches for 3D object detection~\cite{Bevfusion-Liang}. 

% Despite some of these developments, a significant challenge remains when it comes to handling corrupted data,

It is important to mention that the \textit{fusion steps} described above were used in many multimodal fusion methods as stand-alone steps or as one of the intermediate processes in the fusion step. There are also \textbf{\textit{other}} architectures worth mentioning, such as the transformer-based fusion step in Transfusion~\cite{transfusion} or probability voting step in CLOCs~\cite{clocs}. Moreover, some methods use a step architecture that resembles some of the fusion steps  but is hard to classify as one category, such as RoIFusion~\cite{roifusion}. For the sake of this research, we focus on the ones described in depth in this section, since they are used in the best performing methods. 

% or MMF~\cite{mmf}
% \subsection{Summary}

% \subsection{Final note}
% Through presented testing and analysis, we have observed promising results, suggesting that our novel fusion steps have the potential to significantly improve the fusion performance and address the challenges associated with corrupted data.

\section{Methods}
\label{sec:methods}

\textit{Fusion steps} presented in Figure \ref{sec:related} often struggle to maintain the same level of performance on corrupted data such as sensor misalignment, lower resolution, or missing points, as they do on correct data. In light of this, we have proposed a novel fusion step that enhances the robustness and reliability of the fusion process, even in the presence of corrupted data. 

% We drew the inspiration from previously developed fusion steps, alternating them to make them more robust and combining them together to achieve better performance.

\begin{figure*}[!t]
  \begin{center}
    \includegraphics[width=.5\textwidth]{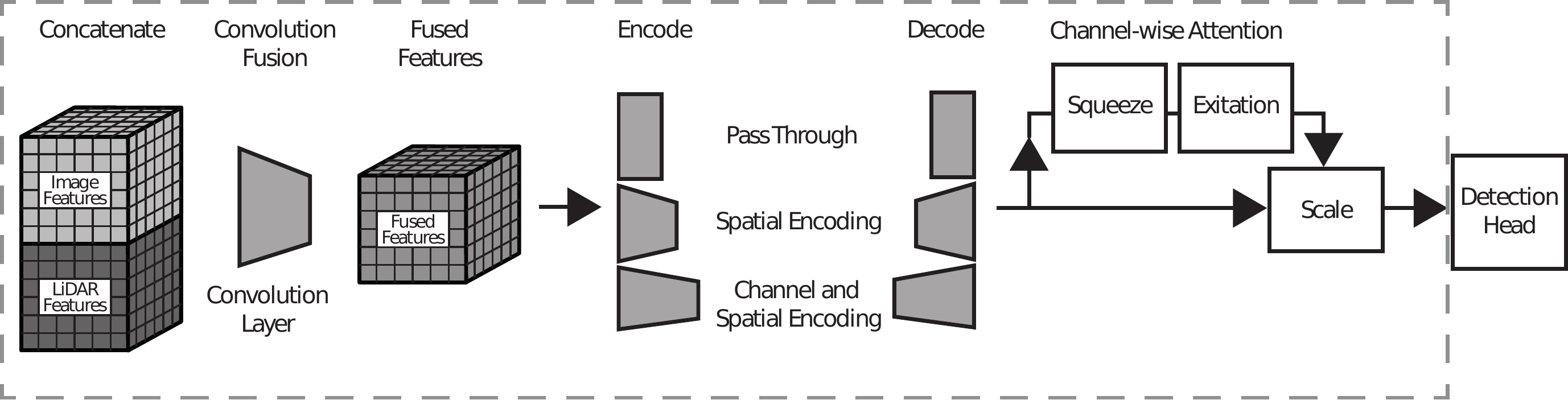}
  \end{center}
  \caption{\textbf{Our method}: the convolution fusion step, followed by encoder-decoder with SE-block.}
  \label{fig:fusion_step_conv_encodedecode_w_se}
\end{figure*}

Our fusion step, shown in Figure \ref{fig:fusion_step_conv_encodedecode_w_se}, drew inspiration from previously developed fusion steps enhancing their robustness by alternating them and combining them together, leading to improved overall performance. 
%Our fusion step achieved high results, even if the camera and LiDAR are misaligned, the LiDAR data has low resolution, is corrupted or partially missing (up to heavy sensor failure).

% specifically focus on these issues. By incorporating innovative techniques and algorithms, our proposed fusion steps

% We propose a novel approach to fuse the camera and LiDAR information in a 3D object detection model. We focus on developing the \textit{fusion step} that will still work well even if data is partially missing (up to even heavy sensor failure) or being misaligned. 
%We explore these scenarios in detail in \cref{sec:experiments}.

%Focused on our methods to be easily understandable

% \subsection{Advanced Encoder-Decoder with SE-block for robust sensor fusion}

% Our proposed novel fusion step, shown in \cref{fig:fusion_step_conv_encodedecode_w_se}, can operate on corrupted LiDAR data. We focus on developing the \textit{fusion step} that will ensure high performance . 

% Thus, multiple instances of an object feature will be connected. 

The fusion step starts with convolution, followed by an encoder-decoder structure and Squeeze-and-Excitation (SE) block. The encoder architecture contains three branches that work in parallel. In the first branch, the fusion features are \textit{passed through} (skip-connection) through a one-layer encoding step. The second \textit{spatial encoding} branch reduces the spatial dimension and upscales to the appropriate dimensions after the decoder layers. The original $200 \times 200$ feature space is reduced to $100 \times 100$ and back again. The third branch performs the same spatial encoding, but also, a channel encoding, where the channel space is funneled to half the original size.

% Next, the features are sent to \textbf{\textit{SE}} block, proposed by Hu et al.~\cite{squeeze-and-excitation} and used in SOTA fusion approaches~\cite{Bevfusion-Liang}. 

% The \textit{squeeze} step makes a global average pooling to aggregate features in order to create channel-wise descriptors. The \textit{excitation} step then uses fully connected layers to produce channel-wise activations that are applied to the map of features. The output of the SE block can then be used to scale the convolutional features, according to information value. Thus, essential interdependencies can be enhanced. 

Next, the SE block attentively scales the relationships between the fused feature channels. The \textit{squeeze} and \textit{excitation} operations enhanced the important information by modeling the channel inter-dependencies through an adaptive average pooling operation, establishing and enhancing the relationships between channels as they might otherwise be lost in the channel-reducing convolution fusion process.

As we emphasized before, our method was carefully designed to address the issues caused by lower sensor resolution, missing LiDAR data or sensor misalignment, but we also want to ensure that our method performs on par with other SOTA methods when data is not corrupted. Therefore, in our encoded-decoder block, the first branch is an information pass-through that allows the model to operate with high performance in a non-misaligned scenario. The other two branches account for sensor-to-sensor misalignment. 

The \textit{spatial encoding branch} facilitates the association of spatial-neighboring features from the two data streams, as they are likely to represent the same object with slight spatial misalignment. The \textit{channel and spatial encoding} branch includes the encoding which is added to account for misalignment in the channel space (between features in the LiDAR and camera feature spaces in the channel direction) since object features can be represented differently in the two feature modalities.

At this stage, the input to the \textit{SE-block} includes features of the same scene encoded in three different ways by each of the encoder-decoder branches. The \textit{SE-block} associates the feature spaces between the branches, ahead of the detection head, leveraging different representations of the object to elevate the overall performance of the 3D object detection model.

\section{Experimental setup}
% \textcolor{red}{1-2 sentences of intro}

This section describes metrics and the experimental setup used for evaluation. Through the experiments, we examine real-world scenarios often occurring in the robotics platforms, when understanding the environment is hindered due to partial sensor failure, sensor misalignment, or lower sensor resolution. 

Our experiments are divided into two main parts. In the first part, we test the robustness of different SOTA methods for 3D object detection and see what performance decrease we can expect to choose the most robust method. We focus on fusion methods, but also test camera and LiDAR-only methods for reference. 
In the second part, we evaluate our proposed fusion step when there is sensor misalignment and with the best performing method from the first part as a baseline by replacing their fusion step with ours.

\subsection{Metrics}
\label{subsec:metric}
Average Precision (\textit{AP}) is defined as the area under the precision-recall curve, $P(r)$. Where \textit{mAP} is the class-wise mean of \textit{AP}. Here, $N$ is the number of classes, and $AP_k$ is the average precision for class $k$, see Equation \ref{eq:mAP}.

\begin{equation}\label{eq:mAP}
        AP = \int P(r)dr \;\;\; mAP = \frac{1}{N}\sum^{k=N}_{k=1} AP_k
\end{equation}

% The definition on Average Precision \textit{AP} is in practice based on the well-established PASCAL VOC~\cite{pascal} definition.

The \textit{KITTI}~\cite{KITTI} metrics require a $70\%$ Intersection over Union (\textit{IoU}) for cars (moderate difficulty) with a minimum bounding box height of 25 pixels and a maximum truncation of $30\%$. The \textit{NuScenes}~\cite{NuScenes} deviates from the definition of a match. It is determined by thresholding the 2D center distance on the ground plane instead of using the \textit{IoU}. This results in \textit{mAP} score being up to \textbf{$2 \times$} higher on \textit{KITTI} than \textit{NuScenes} for the same methods. The mean Average Precision \textit{mAP} is then calculated by threshold averaging, based on distance $\mathbb{D}=\{0.5, 1, 2, 4\}$ in meter. We also used NuScenes Detection Score \textit{NDS} Equation \ref{eq:NDS}, here $\mathbb{TP}$ is a set of five true positive metrics described in detail in~\cite{NuScenes}.

\begin{equation}\label{eq:NDS}
    NDS = \frac{1}{10}(5 \ mAP + \sum^{}_{mTP \in \mathbb{TP}}(1-min(1,~mTP)))
\end{equation}

% , Average Translation Error (ATE), Average Scale Error (ASE), Average Orientation Error (AOE), Average Velocity Error (AVE), and Average Attribute Error (AAE). The respective mean metric is represented by mTP.

\subsection{LiDAR data corruption}

\begin{figure}[!b]
     \centering
     \begin{subfigure}[b]{0.32\columnwidth}
         \centering
         \includegraphics[width=\textwidth]{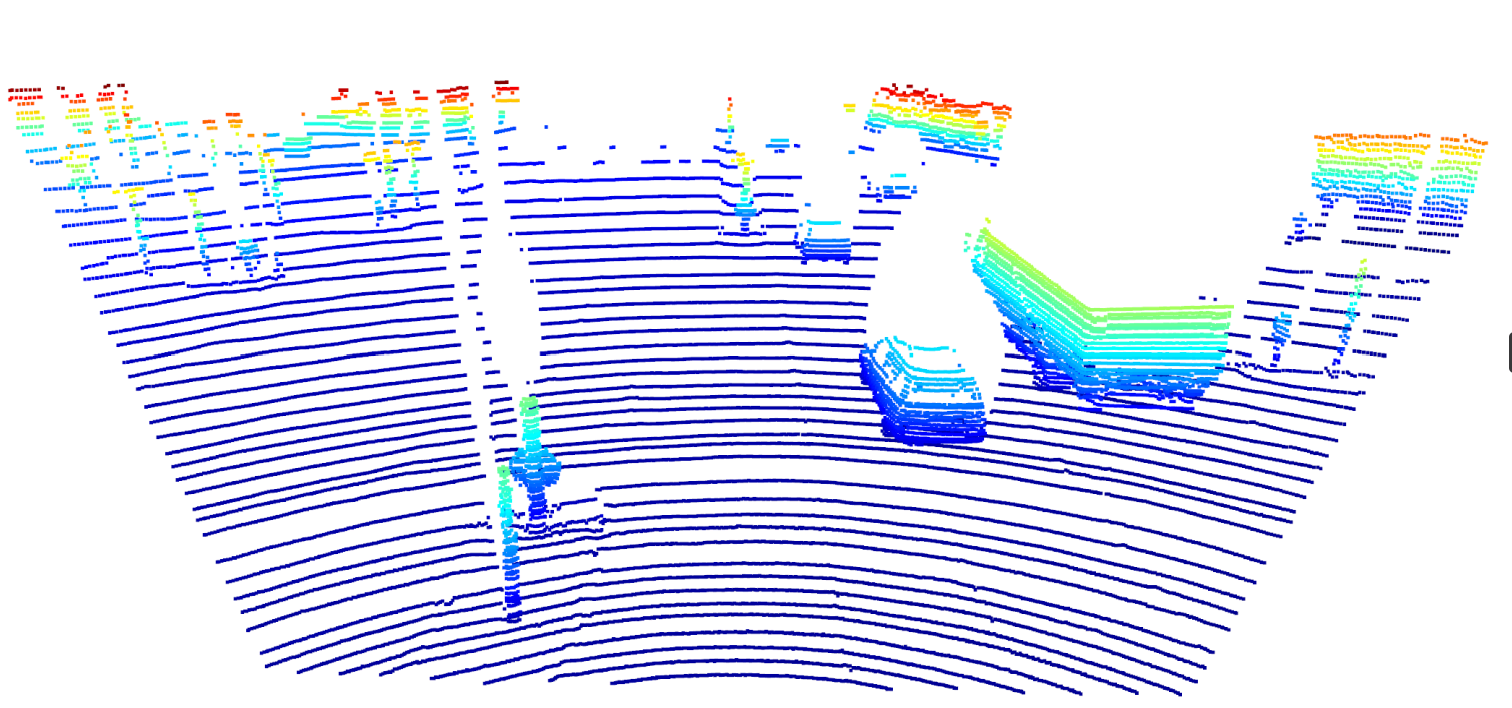}
         \caption{Original LiDAR point cloud.}
         \label{fig:64_layer_kitti}
     \end{subfigure}
     \hfill
     \begin{subfigure}[b]{0.32\columnwidth}
         \centering
         \includegraphics[width=\textwidth]{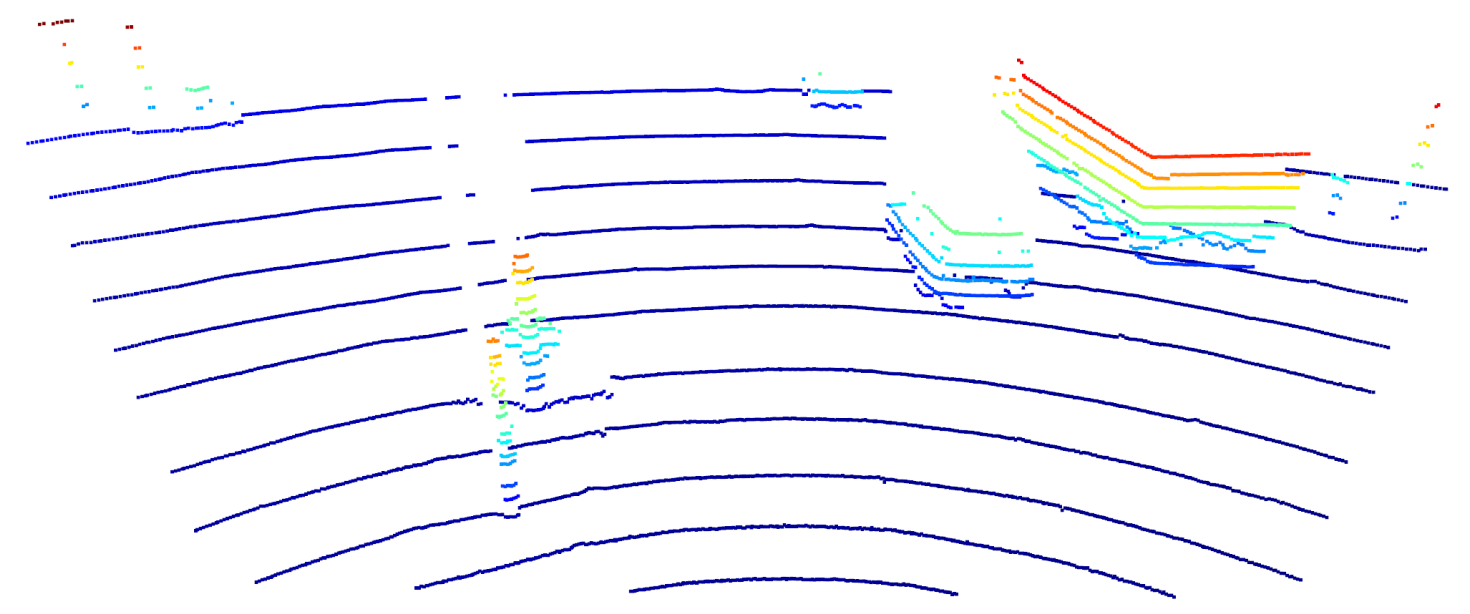}
         \caption{LiDAR point cloud reduced to 16 layers.}
         \label{fig:16_layer_kitti}
     \end{subfigure}
     \hfill
     \begin{subfigure}[b]{0.32\columnwidth}
         \centering
         \includegraphics[width=\textwidth]{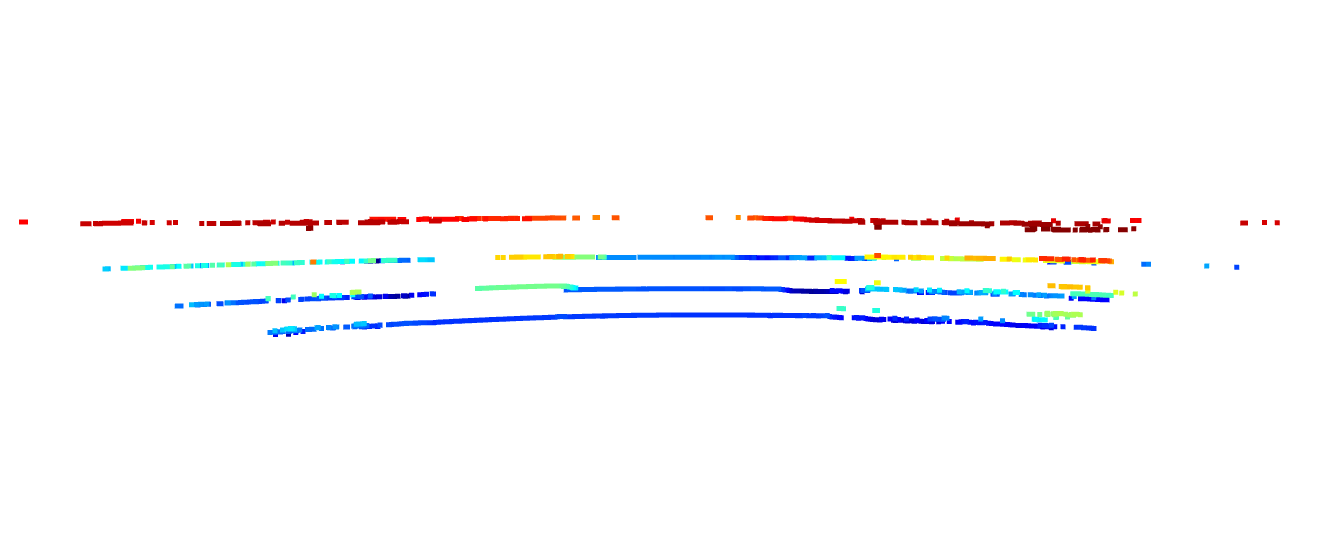}
         \caption{LiDAR point cloud reduction to 4 layers.}
         \label{fig:4_layer_kitti}
     \end{subfigure}
     \hfill
     \begin{subfigure}[t]{0.49\columnwidth}
         \centering
         \includegraphics[width=\textwidth]{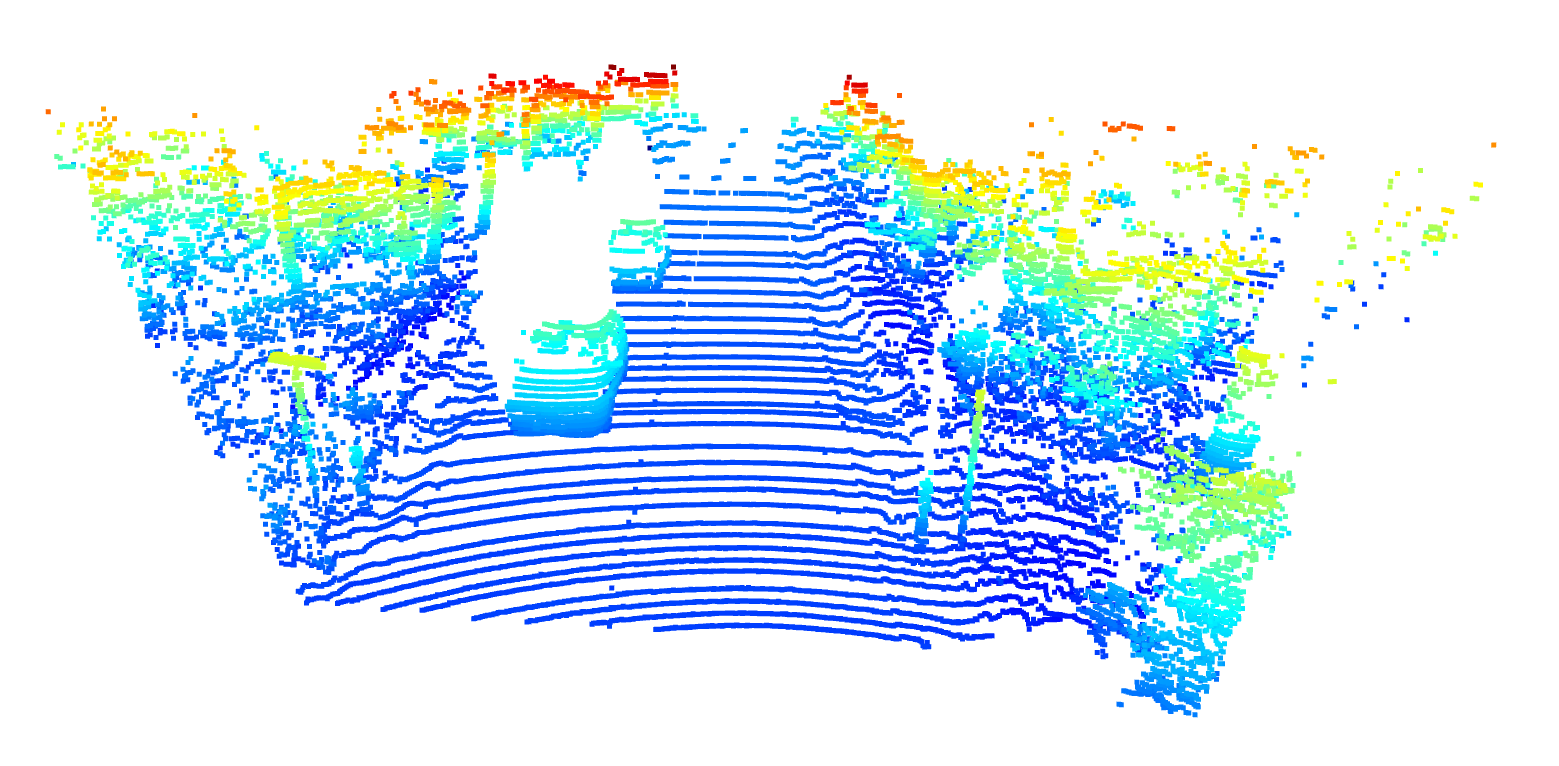}
         \caption{Reference point cloud}
         \label{fig:reducing_lidar_nr_lidar_points_ref}
     \end{subfigure}
     \hfill
     \begin{subfigure}[t]{0.49\columnwidth}
         \centering
         \includegraphics[width=\textwidth]{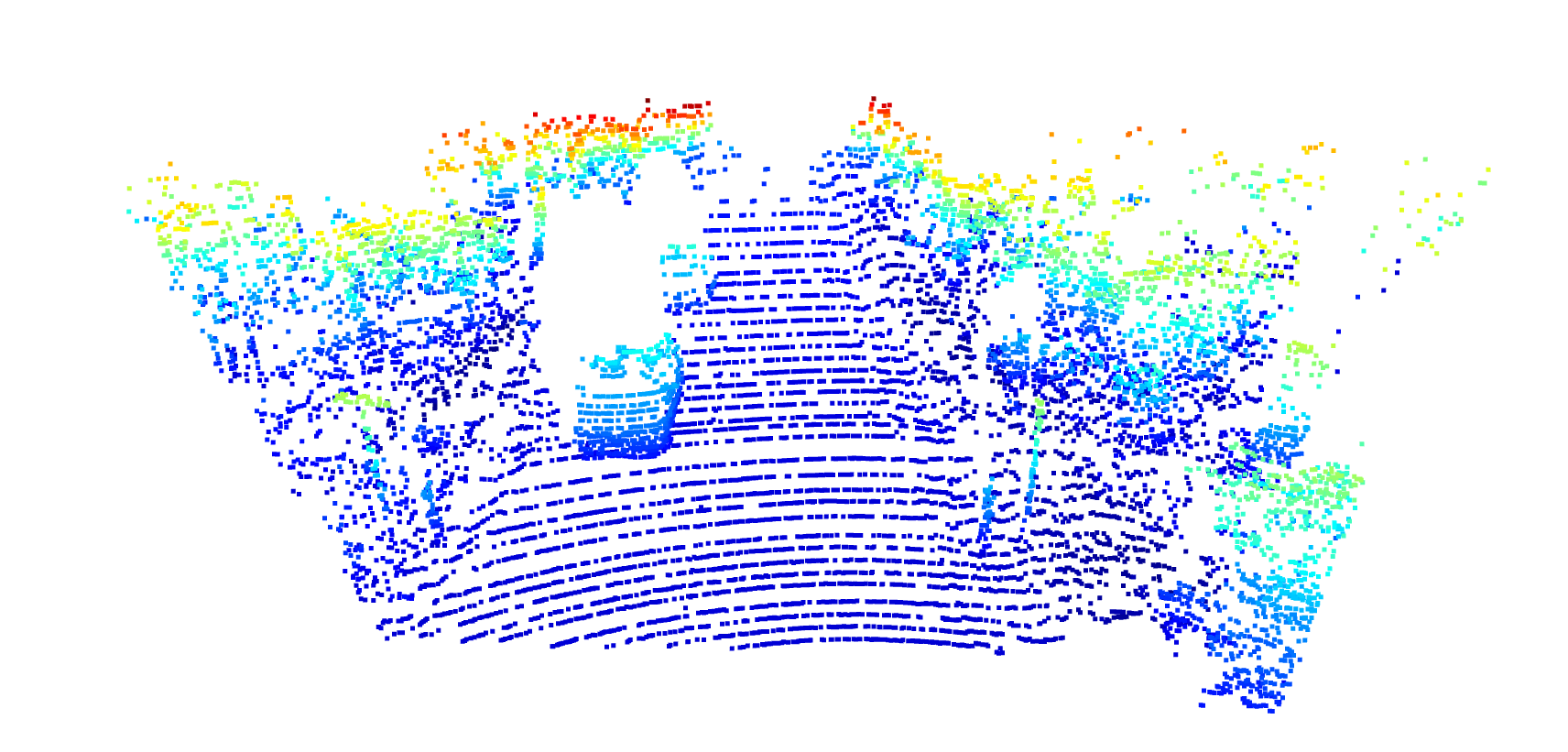}
         \caption{Point cloud data with $50\%$ point reduction.}
         \label{fig:reducing_lidar_nr_lidar_points_ref_50}
     \end{subfigure}
     \caption{Example of LiDAR layer and points reduction.}
     \label{fig:pointlayer_removal}
\end{figure}

Different datasets, smaller mobile robots, or otherwise less expensive autonomous driving setups commonly make use of lower-resolution LiDAR sensors, as in Figure~\ref{fig:16_layer_kitti} or Figure \ref{fig:4_layer_kitti}. Thus, we simulate and test how reducing LiDAR resolution to $16$, $4$, and $1$ layer impacts the 3D object detection model and fusion steps. 

Additionally, a wide variety of disturbances can impact the quality of measurement from a LiDAR sensor. A generic and overarching strategy to highlight how sensitive the multi-modal object detection methods are to the absence of high-quality LiDAR data is to simulate point cloud density reduction. In cases of low-reflection, due to snow, rain, or other environmental effects, the LiDAR reflection beams can be lost and, thus, the point cloud consists of fewer points Figure \ref{fig:reducing_lidar_nr_lidar_points_ref_50}. We simulate this scenario by removing the points on seeded pseudo-random sampling at different ratios of point dropping. 

To avoid this value inflicting any stochastic differences between each experiment, we use a random seed based on the unique sample key. % in the \textit{NuScenes} dataset. 

% While lower resolution can be expected, one can not expect data to be missing in ordered ways, but rather to affect the whole point cloud data in all directions. Therefore, 

\subsection{Camera-LiDAR missaglingment}

To further test the fusion steps with respect to the multi-sensor misalignment problem and misalignment as a result of poor calibration, we propose a misalignment experiment in which the fusion steps will be subject to purposefully added misalignment between the camera and the LiDAR sensor.

%motivated by the slight rotational misalignment of the \textit{LAURA}~\cite{thiel2022integrierte} platform, see \ref{fig:laura_sensor_misal}.

\begin{figure}[!t]
     \centering
     \begin{subfigure}[t]{0.48\columnwidth}
         \centering
         \includegraphics[width=\textwidth]{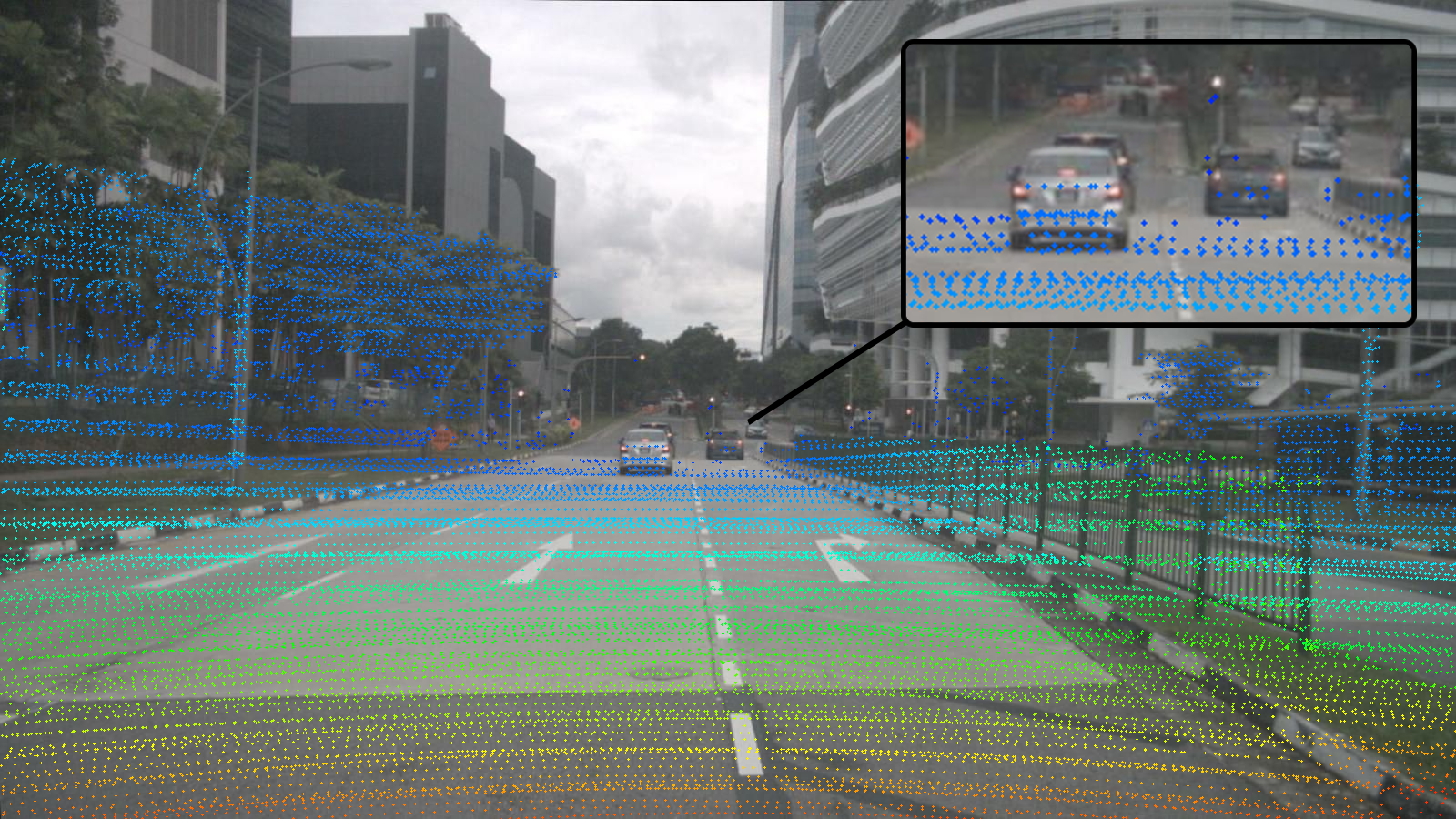}
         \caption{Sensor misalignment example. Reference image.}
         \label{fig:nusc_misalg_ref}
     \end{subfigure}
     \hfill
     \begin{subfigure}[t]{0.48\columnwidth}
         \centering
         \includegraphics[width=\textwidth]{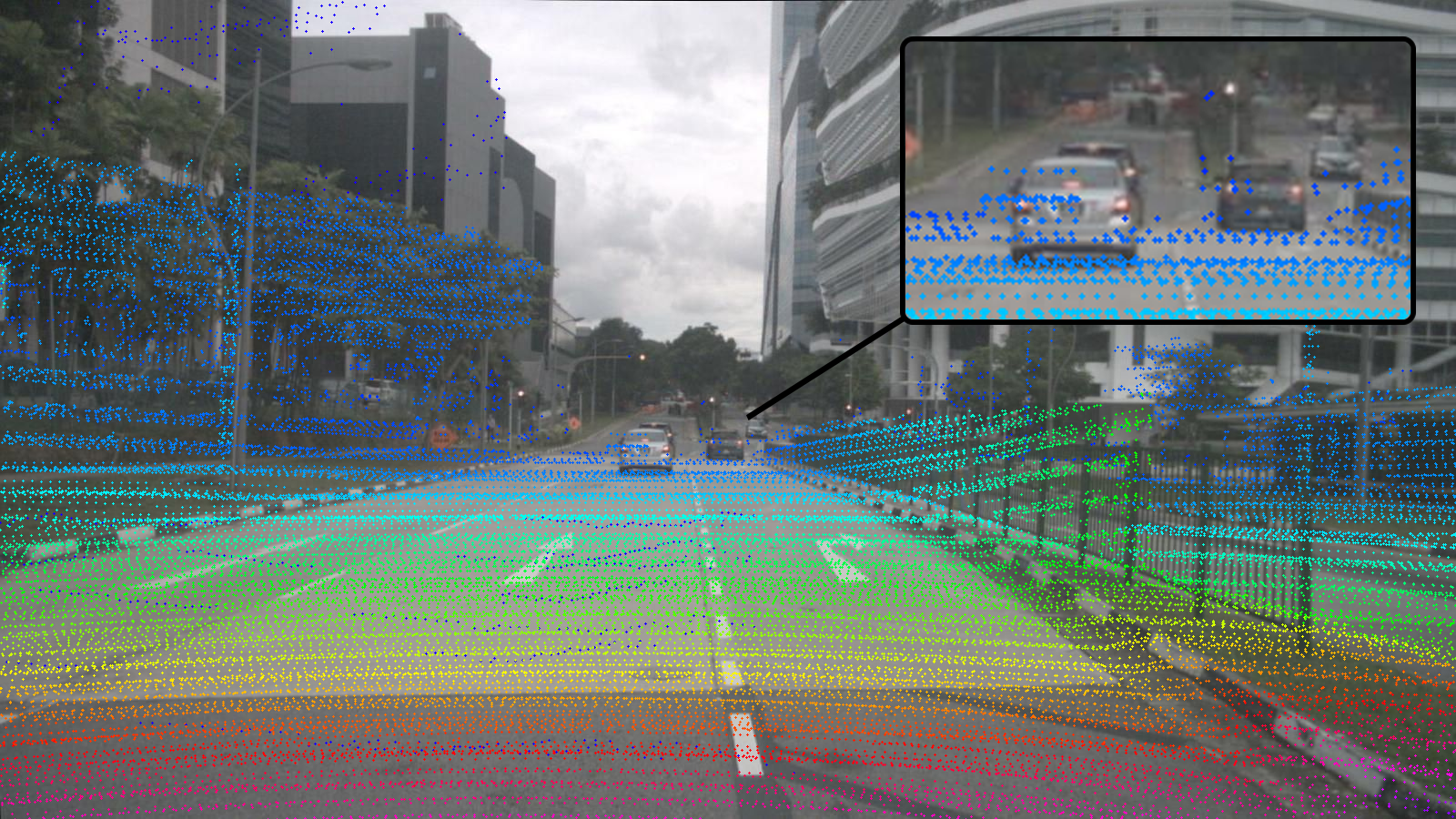}
         \caption{Sensor misalignment example. Translational noise in x,y, and z directions.}
         \label{fig:nusc_misalg_xyz}
     \end{subfigure}

     \begin{subfigure}[t]{0.48\columnwidth}
         \centering
         \includegraphics[width=\textwidth]{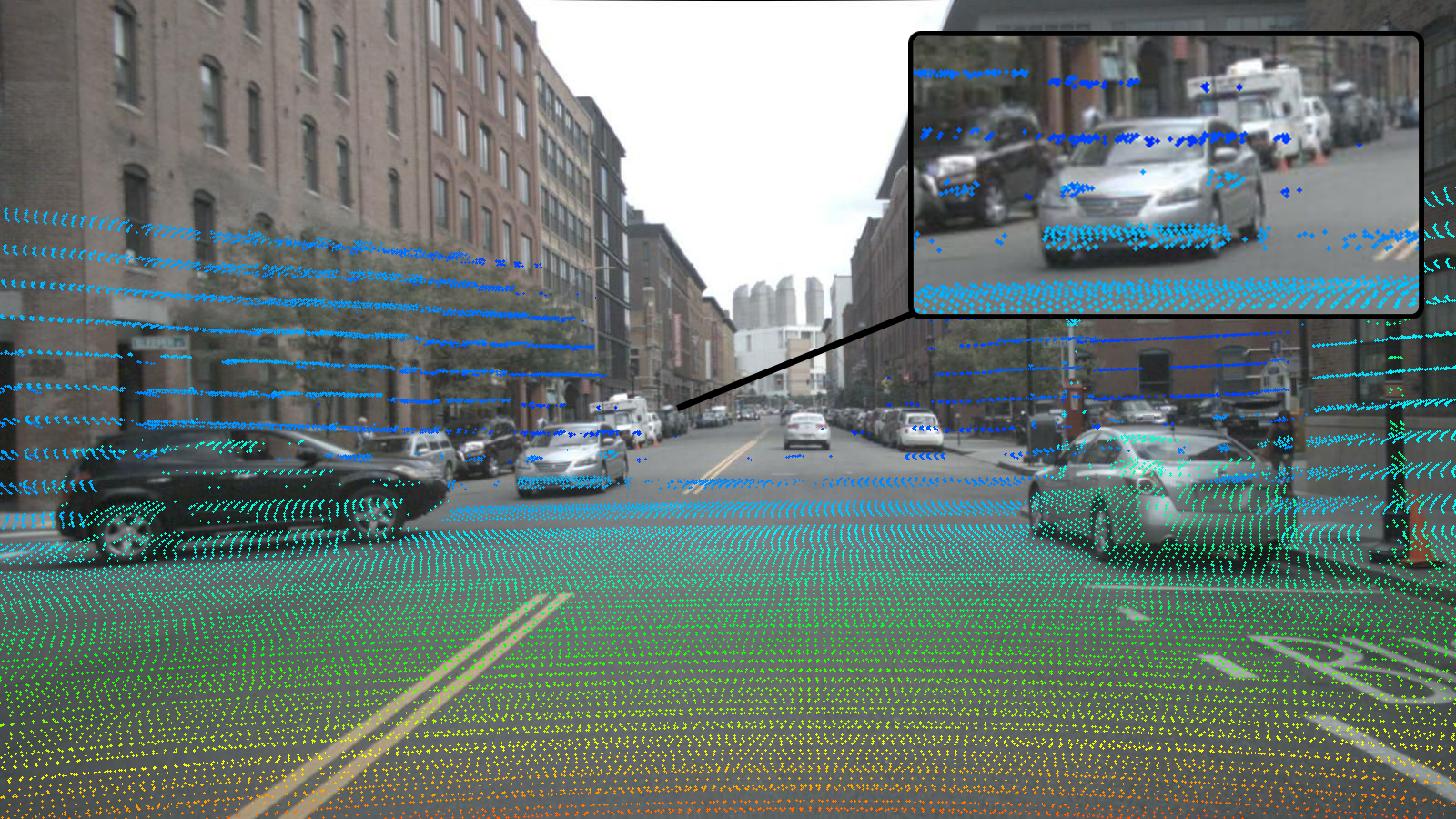}
         \caption{Sensor misalignment example. Reference image.}
         \label{fig:nusc_misalg_ref_2}
     \end{subfigure}
     \hfill
     \begin{subfigure}[t]{0.48\columnwidth}
         \centering
         \includegraphics[width=\textwidth]{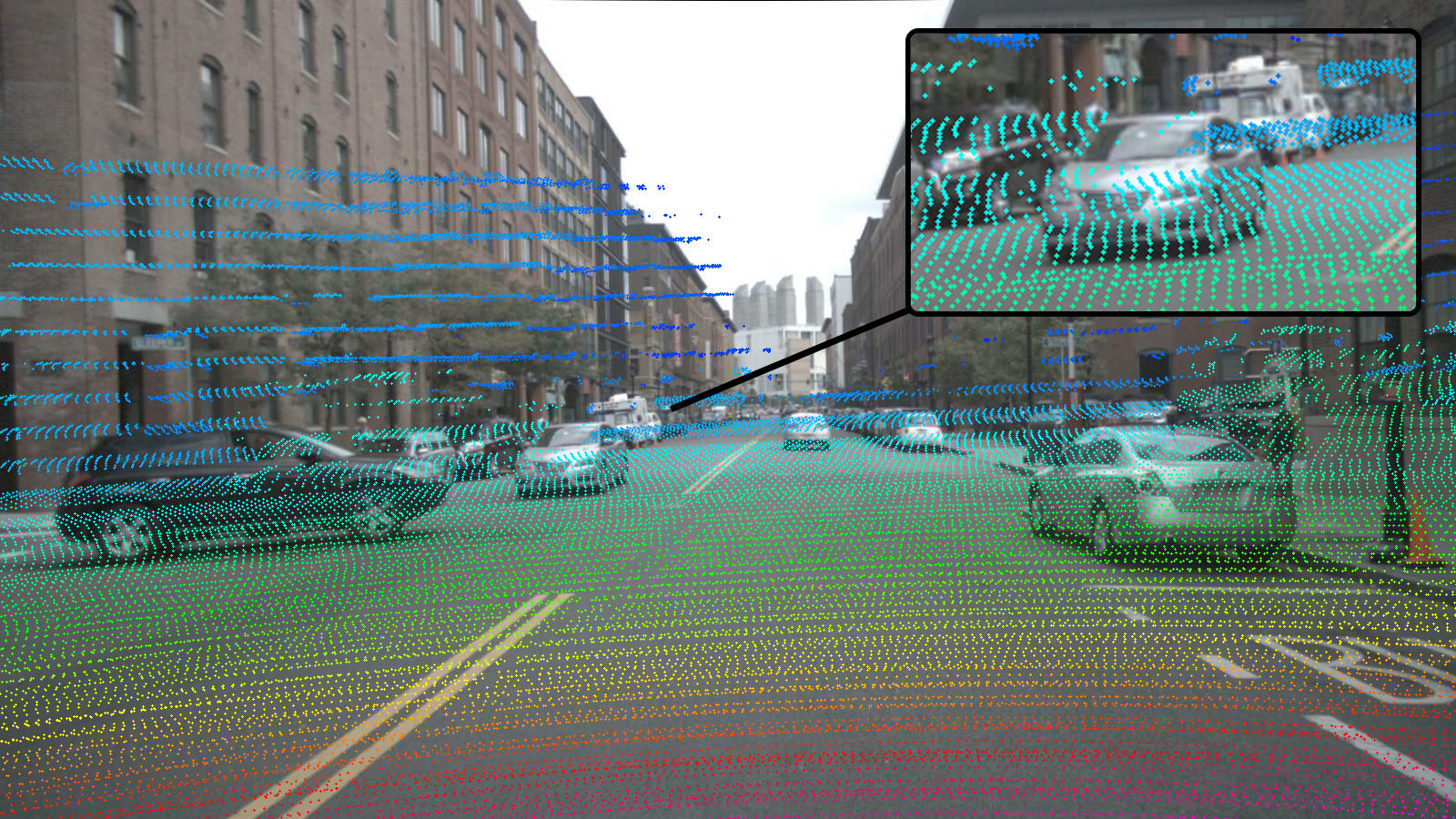}
         \caption{Sensor misalignment example. Rotational noise in roll, pitch, and yaw angles.}
         \label{fig:nusc_misalg_all_rots}
     \end{subfigure}
     \caption{Visualization of sensors misalignment problem.}
\end{figure}

In addition to the tests proposed by \cite{transfusion}, which only handles translation misalignment, our experiments also include rotational misalignment and the combination of both, a commonly occurring problem in the robotics field. This is achieved by adding random noise to the transfer matrices between the respective camera frames, and the joint reference frame, resulting in a shift between the point cloud and camera, as shown in Figure \ref{fig:nusc_misalg_ref} and \ref{fig:nusc_misalg_xyz}. Thus, in the joint feature space, the soon-to-be-fused features are spatially misaligned, and the fusion step must be performed in a way that respects any such misalignment.

% We perform this experiment by adding uniform noise within a fixed bound, for example, $N \in [-10cm, 10cm)$. 

% While new random noise is generated at each sample, these random numbers are based on the sample-unique key. Thus, the noise retains the same pseudo-random values in every experiment, as long as the dataset remains unchanged. 

The experiment is performed on a series of translation and rotation misalignments. In the translation case, noise is added to all three directions simultaneously, $x,y,z$.  In the rotational experiment, noise is added at the same time to $roll, pitch$, and $yaw$.

\section{Experiments}
\label{sec:experiments}
%%%%%%%%%%%%%%%%%%%%%%%%%%%%%%%%%%%%%%%%%%%%%%%%%%%%%%%%%%%%%%%%%%%%%%%%%%%%%%%%%%%%%%%%%%%%%%%%%%%%%%%%
% table 1 nUscenes

\begin{table}[h!]
\caption[Result from the LiDAR layer removal]{Result from the LiDAR layer removal and LiDAR point reduction on NuScences~\cite{NuScenes}.} 
\label{tab:lidar_layer_reduction_results}
\vspace{.5em}
\centering
\begin{adjustbox}{max width=\columnwidth}
\begin{tabular}{ll lll l} %*{3}{S[table-format=2.3]}
\multicolumn{3}{c}{} & \multicolumn{3}{c}{\textbf{Scores on NuScenes}~\cite{NuScenes}} \\
%\cmidrule{0-1}
\cmidrule{4-6}
\textbf{Method} & \textbf{Defect} & \textbf{} & \multicolumn{1}{c}{$mAP$} & \multicolumn{1}{c}{$NDS$} & \multicolumn{1}{c}{$\Delta mAP$} \\
\midrule

BEVFusion-Liang~\cite{Bevfusion-Liang}                   & \textit{Layer} & 32 & 54.01 & 60.66 &   \\
with PointPillars~\cite{PointPillars} configuration      & \textit{removal} & 16 & 47.52 & 56.66 & -12.0\%  \\
Fusion type:\textit{ Deep-fusion}                        & & 4 & 42.10 & 52.80 & -22.1\%  \\
Fusion step:\textit{ Convolution with}          & & 1 & 15.23 & 34.75 & -71.8\%  \\

\cmidrule{3-6}

\textit{SE-block}                                                       & \textit{Points} & 100\% & 54.01 & 60.66 &  \\
                                                         & \textit{reduction} & 90\% & 53.83 & 60.53 & -0.3\% \\
                                                         & & 80\% & 53.58 & 60.40 & -0.8\% \\
                                                         & & 50\% & 51.22 & 58.87 & -5.2\% \\
\midrule
% BEVFusion-MIT~\cite{BEVFusion}          & \textit{Layer} & 32 & 68.57 & 71.46 &   \\
% Fusion type:\textit{ Deep-fusion}       & \textit{reduction} & 16 & 57.00 & 64.45 & -16.9\%  \\
% Fusion step:\textit{ Convolution }  & & 4 & 47.80 & 58.22 & -30.3\%  \\
% \textit{with encoder-decoder}                                        & & 1 & 12.17 & 32.34 & -82.3\%  \\
%                                         & \textit{Points} & 100\% & 68.57 & 71.46 &   \\
%                                         & \textit{removal} & 90\% & 68.22 & 71.21 & -0.5\%  \\
%                                         & & 80\% & 67.79 & 70.94 & -1.1\%  \\
%                                         & & 50\% & 65.28 & 69.26 & -4.8\%  \\
% \midrule
TransFusion~\cite{transfusion}          & \textit{Layer} & 32 & 58.95 & 54.27 &   \\
Fusion type:\textit{ Deep-fusion} & \textit{removal} & 16 & 42.40 & 45.24 & -28.1\%  \\
Fusion step:\textit{ Transformers based}        & & 4 & 27.06 & 36.43 & -54.1\%  \\
\textit{on object queries}                                        & & 1 & 02.22 & 11.54 & -96.2\%  \\
\cmidrule{3-6}

                                        & \textit{Points} & 100\% & 58.95 & 54.27 &  \\
                                        & \textit{reduction} & 90\% & 58.48 & 54.05 & -0.9\%  \\
                                        & & 80\% & 57.83 & 53.69 & -1.8\%  \\
                                        & & 50\% & 53.79 & 51.51 & -8.7\%  \\
\midrule
PointPillars~\cite{PointPillars}      & \textit{Layer} & 32 & 39.71 & 53.15 &   \\
\textit{Single modal: LiDAR only}       & \textit{removal} & 16 & 28.64 & 46.59 & -27.9\%  \\
                                        & & 4 & 15.61 & 38.35 & -60.7\%  \\
                                        & & 1 & 0.64 & 11.61 & -98.4\%  \\
\cmidrule{3-6}
                                    
                                        & \textit{Points} & 100\% & 39.71 & 53.15 &  \\
                                        & \textit{reduction} & 90\% & 39.40 & 52.98 & -0.8\%  \\
                                        & & 80\% & 39.03 & 52.72 & -1.9\% \\
                                        & & 50\% & 36.39 & 51.12 & -8.8\%  \\
\midrule
FCOS3D~\cite{FCOS3D}                 &\textit{Layer removal}  & 32, 16, 4, 1 & 29.80 & 37.74 & -    \\
\cmidrule{3-6}

Single modal:       &\textit{Points reduction} &$100\%,90\%,$   & 29.80 & 37.74 & -   \\
 \textit{RGB camera only (no effected by Lidar)} &   & 80\%, 50\% &  &  &   \\
\midrule 
\multicolumn{3}{l}{$^{1}${$\Delta mAP$ indicate the percentage decrease in }} \\
\multicolumn{3}{l}{$mAP$ in relation to the non-reduced baseline} \\

\end{tabular}
\end{adjustbox}
\end{table}

%%%%%%%%%%%%%%%%%%%%%%%%%%%%%%%%%%%%%%%%%%%%%%%%%%%%%%%%%%%%%%%%%%%%%%%%%%%%%%%%%%%%%%%%%%%%%%%%%%%%%%%%%%%%%%%%%

%%%%%%%%%%%%%%%%%%%%%%%%%%%%%%%%%%%%%%%%%%%%%%%%%%%%%%%%%%%%%%%%%%%%%%%%%%%%%%%%%%%%%%%%%%%%%%%%%%%%%%%%%%%%%%%%%
% KITTI experiments 

\begin{table}[h!]
\caption[Result from the LiDAR layer removal]{Result from the LiDAR layer removal and LiDAR point
reduction on KITTI~\cite{KITTI}.} 
\label{tab:lidar_layer_reduction_results_kitti}
\vspace{.5em}
\centering
\begin{adjustbox}{max width=\columnwidth}
\begin{tabular}{ll lll l} %*{3}{S[table-format=2.3]}

\multicolumn{3}{c}{} & \multicolumn{3}{c}{\textbf{Scores on KITTI}~\cite{KITTI}} \\
% \multicolumn{2}{c}{} & \multicolumn{3}{c}{\scriptsize{(AP40@Moderate)}} \\
\cmidrule{4-6}

\textbf{Metod} & \textbf{Defect} &  & \multicolumn{1}{c}{$mAP_{3D}$} & \multicolumn{1}{c}{$mAP_{bbox}$} & \multicolumn{1}{c}{$\Delta mAP_{bbox}$} \\
\midrule
MVX-Net~\cite{MVXNET}                   & \textit{Layer} & 64 & 62.92 & 75.54 &   \\
Fusion type:\textit{ Early-fusion}      & \textit{removal} & 16 & 43.48 & 56.23 & -30.9\%  \\
Fusion step: \textit{Point-wise}                  & & 4 & 6.04 & 14.62 & -90.4\%  \\
\textit{concatenate}
                                        & & 1 & 0.02 & 0.90 & -99.9\%  \\ 
\cmidrule{3-6}

                                        & \textit{Points} & 100\% & 62.92 & 75.54 &  \\
                                        & \textit{reduction} & 90\% & 62.37 & 75.30 & -0.8\% \\
                                        & & 80\% & 61.48 & 74.41 & -2.3\% \\
                                        & & 50\% & 56.38 & 69.51 & -10.4\% \\ 
\midrule
CLOCs~\cite{clocs}                      & \textit{Layer} & 64 & 69.02 & 81.58 &  \\
Fusion type:\textit{ Late-fusion}       & \textit{removal} & 16 & 46.51 & 63.26 &  -32.6\% \\
Fusion step:\textit{ Object}            & & 4 & 8.04 & 34.18 & -88.4\%  \\
\textit{candidate probability
scoring}                                & & 1 & 1.98 & 11.99 &  -97.1\% \\
\cmidrule{3-6}

Note:\textit{ Calculated from }  
                                        & \textit{Points} & 100\% & 69.02 & 81.58 &  \\
\textit{class-specific models}
                                        & \textit{reduction} & 90\% & 67.88 & 80.15 &  -1.7\% \\
                                        & & 80\% & 66.96 & 78.74 &  -3.0\% \\
                                        & & 50\% & 60.30 & 74.19 & -12.6\% \\
  
\midrule
PointPillars~\cite{PointPillars}        & \textit{Layer} & 64 & 64.36 & 75.43 &   \\
Single modal:\textit{ LiDAR only}       & \textit{removal} & 16 & 47.96 & 65.63 & -25.5\%  \\
                                        & & 4 & 15.22 & 20.07 & -76.4\%  \\
                                        & & 1 & 0.90 & 5.19 & -98.6\%  \\
\cmidrule{3-6}
                                        
                                        & \textit{Points} & 100\% & 64.36 & 75.43 & \\
                                        & \textit{reduction} & 90\% & 63.63 & 75.43 & -1.1\% \\
                                        & & 80\% & 62.11 & 74.51 & -3.5\% \\
                                        & & 50\% & 55.73 & 69.82 & -13.4\% \\
\midrule 
SMOKE~\cite{SMOKE}                      &\textit{Layer removal} & 64, 16, 4, 1 & 3.51 & 52.83 &  -  \\
\cmidrule{3-6}

Single modal:  &\textit{Points reduction} & 100\%, 90\%, & 3.51 & 52.83 &  - \\
 \textit{RGB camera only} &   & 80\%, 50\% &  &  &   \\
  \textit{(not effected by LiDAR)} &   &   &  &  &   \\
\midrule 
\multicolumn{3}{l}{$^{1}${$\Delta mAP$ indicate the percentage decrease in }} \\
\multicolumn{3}{l}{$mAP$ in relation to the non-reduced baseline} \\
\end{tabular}
\end{adjustbox}
\end{table}

Our experiments are divided into two main parts. In the first part, we test the robustness of different SOTA methods for 3D object detection and see what performance decrease we can expect to choose the most robust method.
% Next, we compare the method with the baseline fusion step with our fusion step. 

In the second part, we use the most robust model and test SOTA and our fusion steps against sensor misalignment.

% We show that our fusion step is the most robust for various data corruption and outperforms other approaches used by the state-of-the-art methods. 
% make sure it is true and add the final statement here

\begin{figure*}[t!]
     \centering
     \begin{subfigure}[t]{0.3\textwidth}
         \centering
         \includegraphics[width=\textwidth]{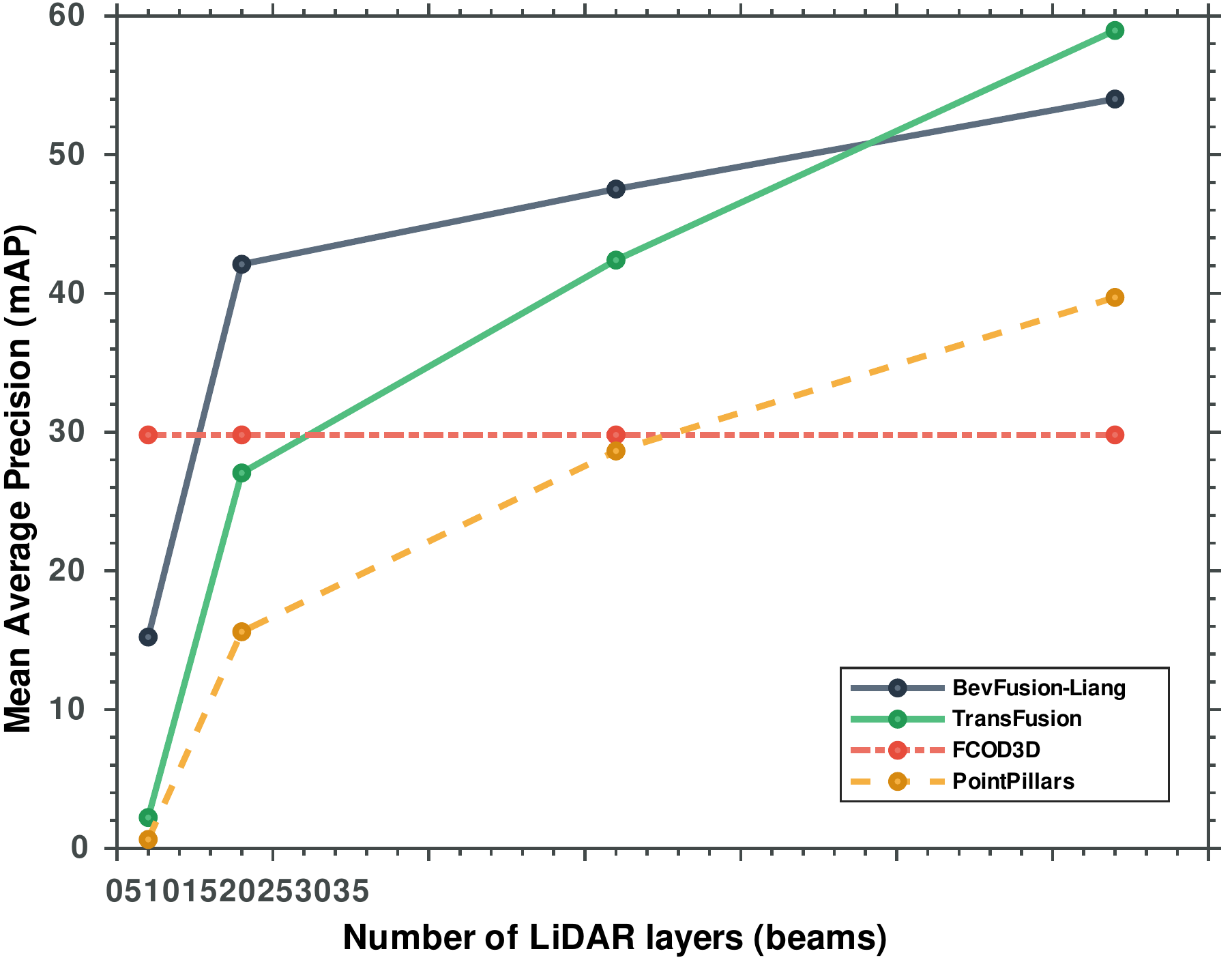}
         \caption{LiDAR \textbf{layer removal} on the NuScenes ($mAP$)}
         \label{fig:reduced_lidar_nusc_map}
     \end{subfigure}
     \hfill
     \begin{subfigure}[t]{0.3\textwidth}
         \centering
         \includegraphics[width=\textwidth]{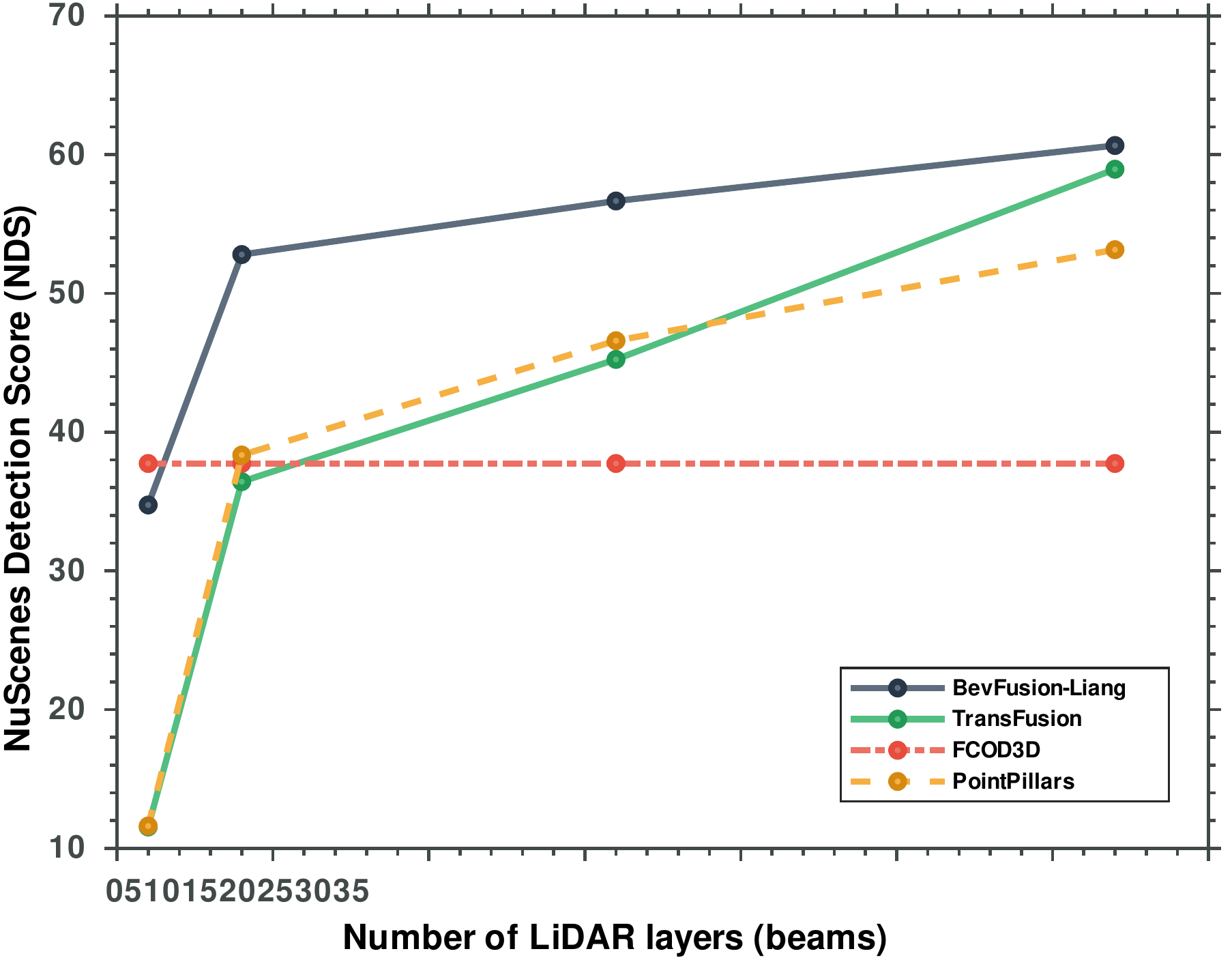}
         \caption{LiDAR \textbf{layer removal} on the NuScenes ($NDS$)}
         \label{fig:reduced_lidar_nusc_nds}
     \end{subfigure}
     \hfill
     \begin{subfigure}[t]{0.3\textwidth}
         \centering
         \includegraphics[width=\textwidth]{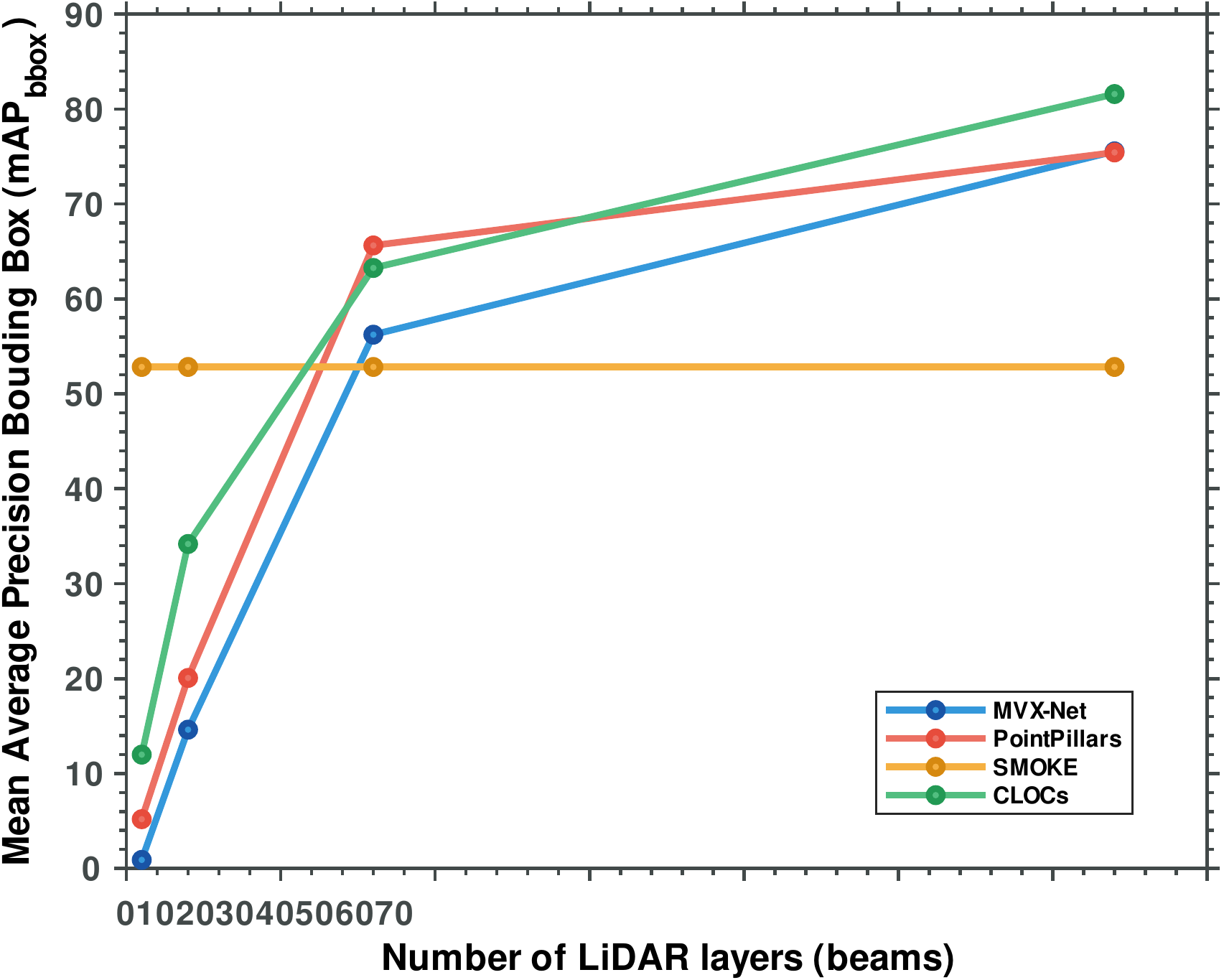}
         \caption{LiDAR \textbf{layer removal} on the KITTI ($mAP_{bbox}$)}
         \label{fig:reduced_lidar_kitti_map_bbox}
     \end{subfigure}
  
     \begin{subfigure}[t]{0.3\textwidth}
         \centering
         \includegraphics[width=\textwidth]{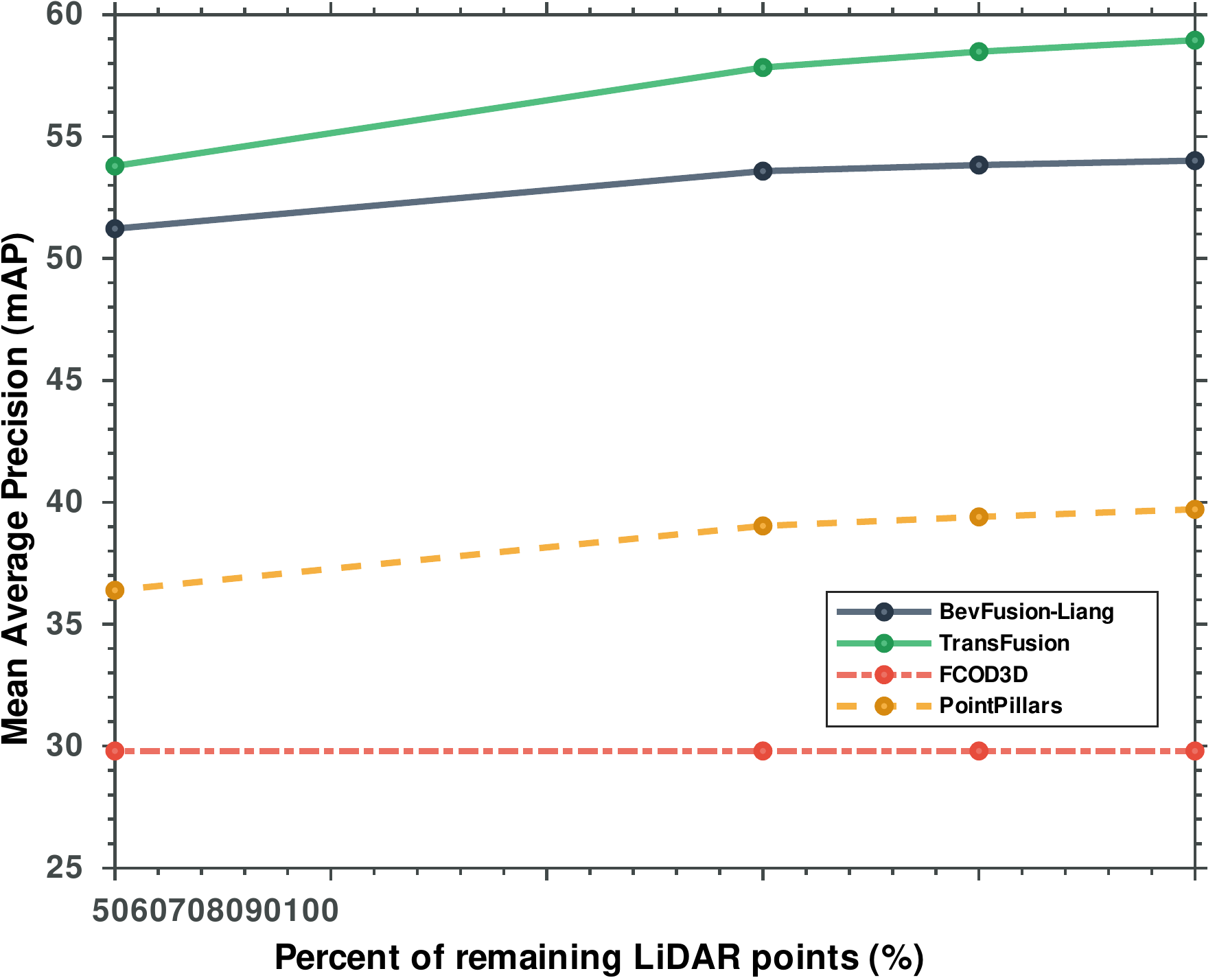}
         \caption{LiDAR \textbf{point number reduction} on the NuScenes ($mAP$)}
         \label{fig:reduced_point_nusc_map}
     \end{subfigure}
     \hfill
     \begin{subfigure}[t]{0.3\textwidth}
         \centering
         \includegraphics[width=\textwidth]{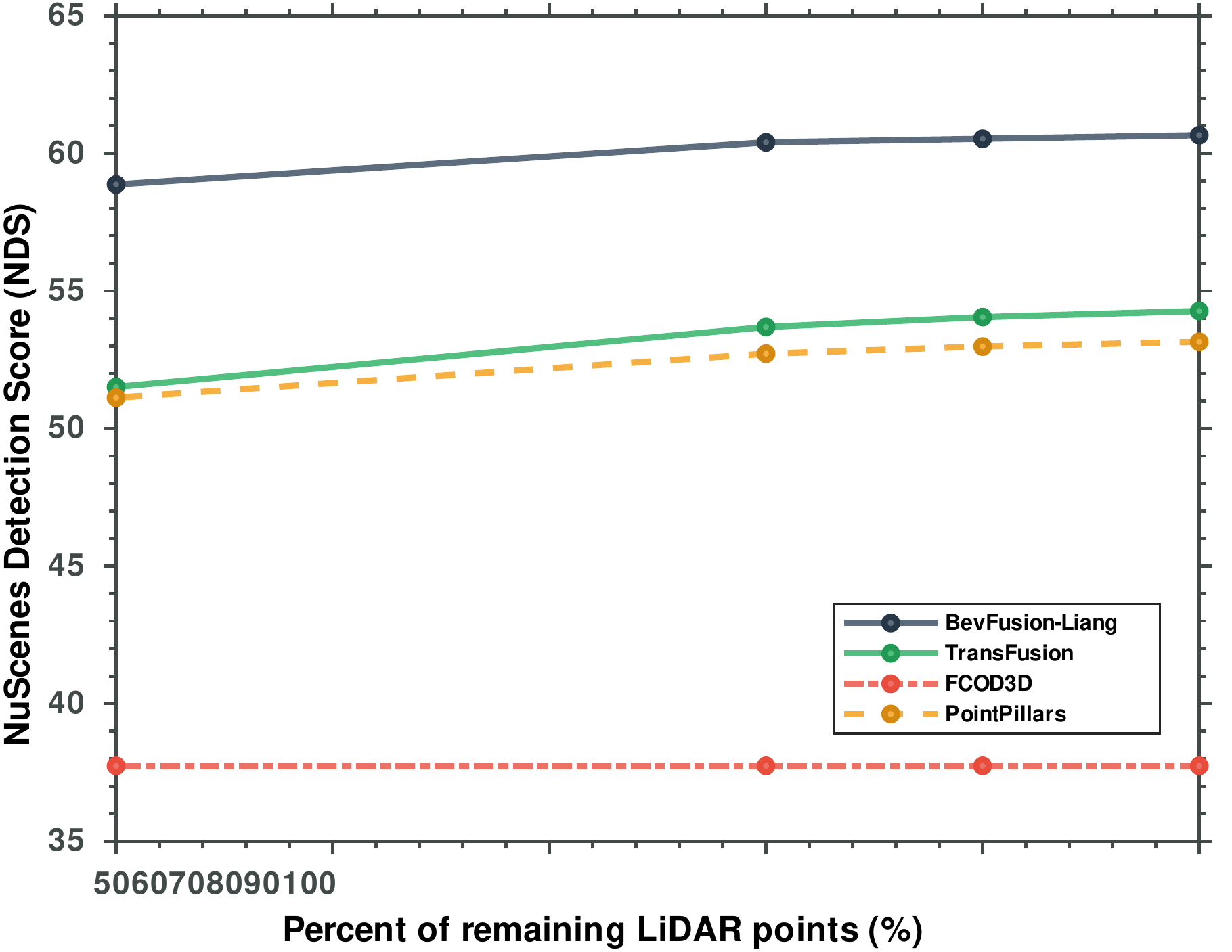}
         \caption{LiDAR \textbf{point number reduction} on the NuScenes ($NDS$)}
         \label{fig:reduced_point_nusc_nds}
     \end{subfigure}
     \hfill
     \begin{subfigure}[t]{0.3\textwidth}
         \centering
         \includegraphics[width=\textwidth]{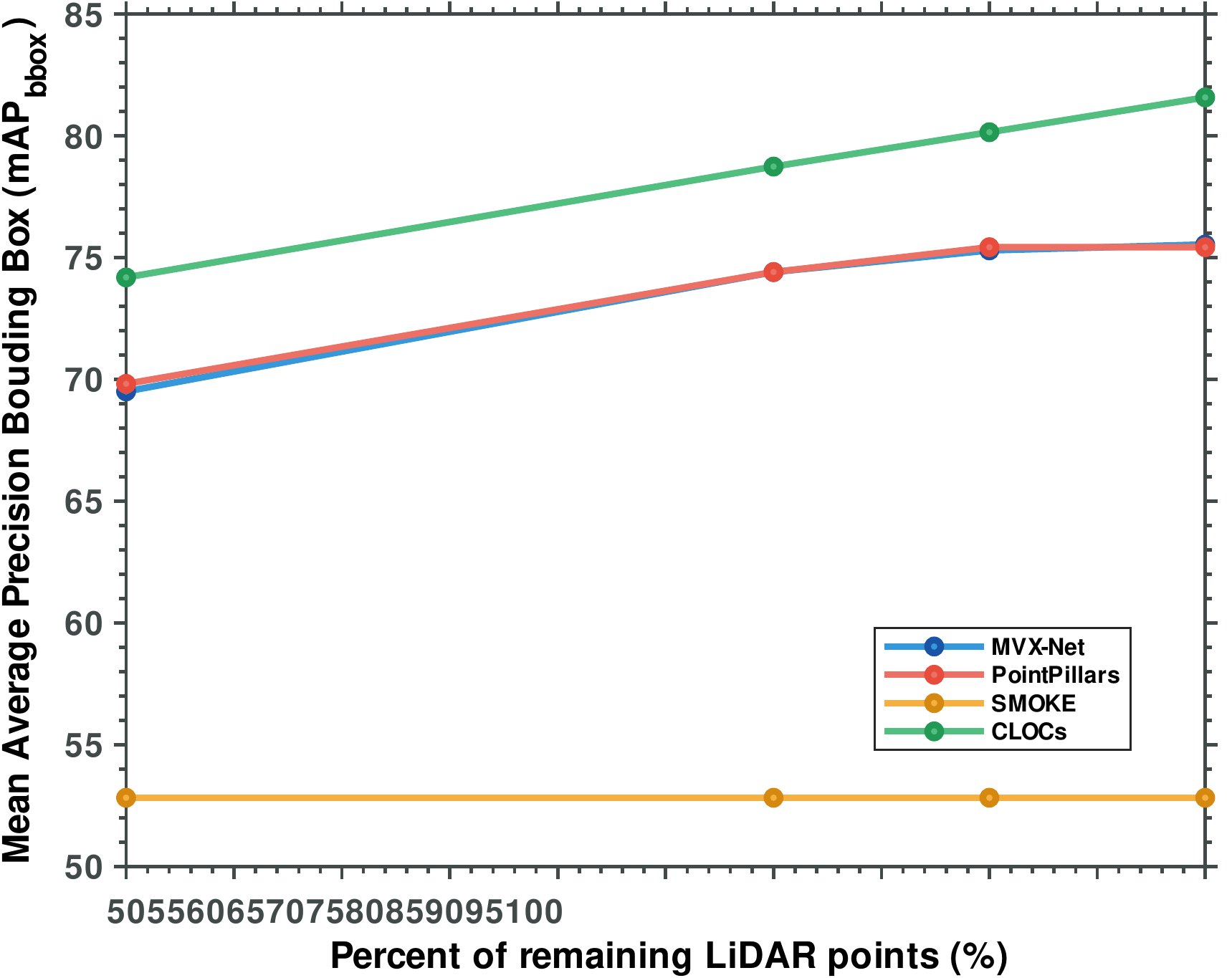}
         \caption{LiDAR \textbf{point number reduction} on the KITTI ($mAP_{bbox}$)}
         \label{fig:reduced_point_kitti_map_bbox}
     \end{subfigure}
     \caption{Comparison of different methods' performance when tested on data with reduced LiDAR layers and reduced point cloud.}
     \label{fig:benchmarkmodels}
\end{figure*}

\subsection{Robustness experiments - layer and LiDAR point removal}
\label{subsec:exp_removal}

In the first two experiments, we evaluate SOTA methods on lower density LiDAR data. In Table \ref{tab:lidar_layer_reduction_results} and \ref{tab:lidar_layer_reduction_results_kitti} the $fusion~type$ section denotes at what level the respective method fuses the data streams using the taxonomy as introduced in Section \ref{sec:related}. The $fusion \; step$ section denotes how that fusion is realized. 

% note that acc on kitti is usually 2x nuscenes --> added in methods 
%In the LiDAR layer data removal experiment, the \textit{KITTI}~\cite{KITTI} and the \textit{NuScenes}~\cite{NuScenes} datasets are modified to simulate lower resolution data. 
In the LiDAR layer removal experiment, each method is evaluated on $16, 4$, and $1$ layered point cloud data on $64$~(\textit{KITTI}) or $32$~(\textit{NuScenes}). 
The results are shown in Table \ref{tab:lidar_layer_reduction_results} and Table \ref{tab:lidar_layer_reduction_results_kitti} and Figure \ref{fig:benchmarkmodels}, highlight how the early-fusion method, MVX-Net~\cite{MVXNET} and the late-fusion method, CLOCS~\cite{clocs} show significant performance drops as the layer number decreased as compared to Transfusion~\cite{transfusion} and BEVFusion-Liang~\cite{Bevfusion-Liang}, deep-fusion methods. The single-modal LiDAR-only PointPillars~\cite{PointPillars} largely follow the decrease in performance. This highlights the fact that presented fusion methods still heavily depend on high-resolution LiDAR data and largely fail to operate independently on the unaffected RGB images when LiDAR data is corrupted. 

%Besides fusion methods, we also tested camera and LiDAR-only methods for reference. 

% The results are presented using \textit{mAP} and \textit{NDS} as described in \cref{subsec:metric}. A percentage decrease in $mAP$ in relation to the non-layer-reduced case is additionally presented.

In the point cloud reduction experiment, we evaluate each method on point clouds randomly reduced to $90\%, 80\%$, and $50\%$ in order to simulate LiDAR performance deviations. We can once again observe how the BEVFusion-Liang method \cite{Bevfusion-Liang} can retain performance as the LiDAR point cloud data is affected negatively. Further, the early fusion method MVX-Net~\cite{MVXNET}, outperforms the LiDAR-only PointPillars~\cite{PointPillars} model in the most extreme case. In this case, the point cloud is affected in an unordered way and fusion methods find a way to associate corresponding point clouds with the image feature, resulting in a less significant performance drop. 

When we compare both deep fusion methods, we can see that BEVFusion-Liang~\cite{Bevfusion-Liang} outperforms Transfusion~\cite{transfusion} on LiDAR layer removal and is slightly worse when it comes to point cloud reduction. Nevertheless, the percentage-wise decrease in performance is much smaller, suggesting it is more robust to various data perturbations.

% After analyzing the results, we chose BEVFusion-Liang~\cite{Bevfusion-Liang} to test different fusion steps architectures since it proved to be the most robust method when expose to data corruption. 

\subsection{Proposed fusion step}
The results above led us to choose BEVFusion-Liang~\cite{Bevfusion-Liang} as the baseline for the evaluation of our proposed fusion step. We provide a short description of their model in the Appendix. 

% since BEVFusion-Liang proved to be the most robust method when exposed to LiDAR layer removal and point cloud reduction. 

%\subsection{Further evaluation of the chosen model}

We benchmark the baseline against a version with our fusion step Section \ref{sec:methods} on the experiments in Section \ref{subsec:exp_removal}.
%LiDAR layer removal and point cloud reduction. 
The results in Table \ref{tab:final_model_lidar_layer_reduction_results} highlight how our fusion step performs very similarly to the baseline BEVFusion~\cite{Bevfusion-Liang} model when it comes to the LiDAR layer removal and point cloud reduction. While our method achieves higher $mAP$ for most LiDAR layer removal scenarios, it performs only slightly worse compared to the baseline on point cloud reduction.

We can also observe that the percentage-wise drop for our fusion step is lower than the baseline, proving that our solution is more robust and less prone to data corruption. 

% Furthermore, our approach surpasses both the baseline and other state-of-the-art fusion step techniques when evaluated for sensor misalignment, exhibiting the lowest performance drop in both easy and hard cases.

%%%%%%%%%%%%%%%%%%%%%%%%%%%%%%%%%%%%%%%%%%%%%%%%%%%%%%%%%%%%%%%%%%%%%%%%%%%%%
\begin{table}[t!]
\caption[Result from the LiDAR layer and point number reduction on our fusion step versus baseline.]{LiDAR layer and point number reduction on our fusion step versus baseline fusion step.} 
\label{tab:final_model_lidar_layer_reduction_results}
\vspace{.5em}
\centering
\begin{adjustbox}{max width=0.95\columnwidth}
\begin{small}\begin{tabular}{ll lll l} %*{2}{S[table-format=2.3]}
\multicolumn{2}{c}{} & \multicolumn{1}{c}{} & \multicolumn{3}{c}{\textbf{Scores on NuScenes}~\cite{NuScenes}} \\
%\cmidrule{0-1}
\cmidrule{4-6}
\textbf{Metod} & \textbf{Defect} & & \multicolumn{1}{c}{$mAP$} & \multicolumn{1}{c}{$NDS$} & \multicolumn{1}{c}{$\Delta mAP$} \\
\midrule

\textbf{Our} fusion step:\textit{ Convolution with}                   & \textit{Layer} & 32 & 51.69 & 56.18 &   \\
\textit{encoder-decoder SE-block}             & \textit{removal} & 16 & 46.38 & 54.22 & -10.2\%  \\
\textit{Augmented}                              & & 4 & \textbf{42.65} & 51.94 & -17.5\%  \\
       & & 1 & \textbf{20.98} & 37.90 & -59.4\%  \\

\cmidrule{3-6}
                                              & \textit{Points} & 100\% & 51.69 & 56.18 &  \\
                                                         & \textit{reduction} & 90\% & 51.65 & 57.45 & -0.01\% \\
                                                         & & 80\% & 51.57 & 57.45 & -0.2\% \\
                                                         & & 50\% & 49.68 & 56.32 & -3.9\% \\
\midrule

Beseline fusion step                  & \textit{Layer} & 32 & 54.01 & 60.66 &   \\
\textit{Convolution with SE-block}~\cite{Bevfusion-Liang} 
      & \textit{removal} & 16 & \textbf{47.52} & 56.66 & -12.0\%  \\
                        & & 4 & 42.10 & 52.80 & -22.1\%  \\
          & & 1 & 15.23 & 34.75 & -71.8\%  \\

\cmidrule{3-6}

                                                      & \textit{Points} & 100\% & 54.01 & 60.66 &  \\
                                                         & \textit{reduction} & 90\% & \textbf{53.83} & 60.53 & -0.3\% \\
                                                         & & 80\% & \textbf{53.58} & 60.40 & -0.8\% \\
                                                         & & 50\% & \textbf{51.22} & 58.87 & -5.2\% \\
\midrule

\end{tabular}\end{small}
\end{adjustbox}
\end{table}
%%%%%%%%%%%%%%%%%%%%%%%%%%%%%%%%%%%%%%%%%%%%%%%%%%%%%%%%%%%%%%%%%%%%%%%%%%%%%%%%%%%%%%%%%%%%%%%%%%%%%%%%%%%%%%%%%%%%%%

%%%%%%%%%%%%%%%%%%%%%%%%%%%%%%%%%%%%%%%%%%%%5%%%%%%%%%%%%%%%%%%%%%%%%%%%%%%%%%%%%%%%%%%%%5
% table fusion step

\begin{table*}[h!]
\caption[Misalignment and fusion steps, optimal training strategy]{3D object detection results from sensor misalignment with different fusion steps experiments on the chosen model~\cite{BEVFusion}.}
\label{tab:fusion_step_mis_alg_opt_results}
\vspace{.5em}
\centering
%\addtolength{\leftskip} {-2cm} % increase (absolute) value if needed
%\addtolength{\rightskip}{-2cm}
\begin{adjustbox}{max width=0.7\textwidth}
\begin{small}\begin{tabular}{ll *{6}l}
\multicolumn{2}{c}{} & \multicolumn{6}{c}{\textbf{Misalignment with max. limits. NuScenes}~\cite{NuScenes}} \\
%\cmidrule{0-1}
\cmidrule{3-8}
\textbf{Fusion Step} & \multicolumn{1}{c}{\textbf{Metric}}  & 
\multicolumn{1}{c}{\textbf{$\mathbf{None}$}} &\multicolumn{1}{c}{$\mathbf{10cm}$} & \multicolumn{1}{c}{$\mathbf{100cm}$} &\multicolumn{1}{c}{$\mathbf{1^{\circ}}$} & \multicolumn{1}{c}{$\mathbf{3^{\circ}}$} &\multicolumn{1}{c}{$\mathbf{10cm \cup 1^{\circ}}$}\\

\midrule

Element-wise add                                  & mAP & 55.33 & 55.27 & 49.52 & 53.47 & \underline{46.73} & 53.37\\
\begin{scriptsize}\textit{as in MVXNet\cite{MVXNET}}\end{scriptsize} 

                                                  & NDS & 55.17 & 52.95 & 48.23 & 51.39 & 45.57 & 51.47 \\  

\midrule

Concatenation                                  & mAP & 47.30 & 47.29 & 41.69 & 46.53 & 37.43 & 39.43\\
\begin{scriptsize}\textit{as in PointPainting\cite{pointpainting}}\end{scriptsize} 

                                               & NDS & 45.34 & 45.70 & 41.55 & 44.80 & 38.36 & 40.32 \\  

\midrule

Fully connected                         & mAP & 54.30 & 53.39 & 45.82 & 51.05 & 43.40 & 49.97\\
\begin{scriptsize}\textit{as in PointFusion\cite{pointfusion}}\end{scriptsize} 

                                        & NDS & 53.87 & 51.18 & 46.19 & 49.53 & 44.22 &  49.17\\  

\midrule

Convolution                     & mAP & 51.06 & 50.54 & 44.71 & 49.79 & 41.70 & 44.69\\
\begin{scriptsize}\textit{as an option in \cite{Bevfusion-Liang} or \cite{BEVFusion}}\end{scriptsize} 

                                & NDS & 49.68 & 48.99 & 44.79 & 48.34 & 42.79 & 43.91 \\

\midrule

Convolution with encoder-decoder      & mAP & 48.72 & 48.50 & 41.73 & 46.75 & 37.74 & 47.20 \\
\begin{scriptsize}\textit{as in BEVFusion-MIT~\cite{BEVFusion}}\end{scriptsize}
                                     & NDS & 49.93 & 48.48 & 43.90 & 46.75 & 41.15 & 47.34\\

\midrule

Convolution with SE-block (\textbf{Baseline})      & mAP  & \textbf{57.20} & \underline{56.23} & 49.18 & \underline{54.51} & 46.40 & \underline{54.08} \\
\begin{scriptsize}\textit{as in BEVFusion-Liang~\cite{Bevfusion-Liang}}\end{scriptsize} 
                              & NDS  & \underline{56.65} & \underline{53.34} & 48.40 & \underline{52.15} & 46.10 & \underline{51.74} \\

\midrule

Convolution with encoder                 & mAP & 55.75 & 55.08 & \underline{49.55} & 53.53 & 44.13 & 52.25 \\
decoder and SE-block (Ours)                    & NDS & 55.24 & 52.32 & \underline{49.00} & 51.68 & \underline{46.11} & 50.59 \\ 

\midrule

Convolution with encoder                 & mAP & \underline{56.34} &  \textbf{57.14} & \textbf{53.92} &  \textbf{55.89} &  \textbf{49.36} &  \textbf{56.15}\\
decoder and SE-block \textit{Augmented} (\textbf{Ours})   & NDS & \textbf{57.02} & \textbf{54.85} &  \textbf{53.31} &  \textbf{54.44} &  \textbf{51.56} &  \textbf{54.33} \\  

\midrule

\multicolumn{3}{l}{$^{1}$\footnotesize{The best results are in \textbf{bold} and second-best are \underline{underlined}.}} \\
\end{tabular}\end{small}
\end{adjustbox}

\end{table*}

\begin{figure*}[!h]
     \centering
     \begin{subfigure}[b]{0.33\textwidth}
         \centering
         \includegraphics[width=\textwidth]{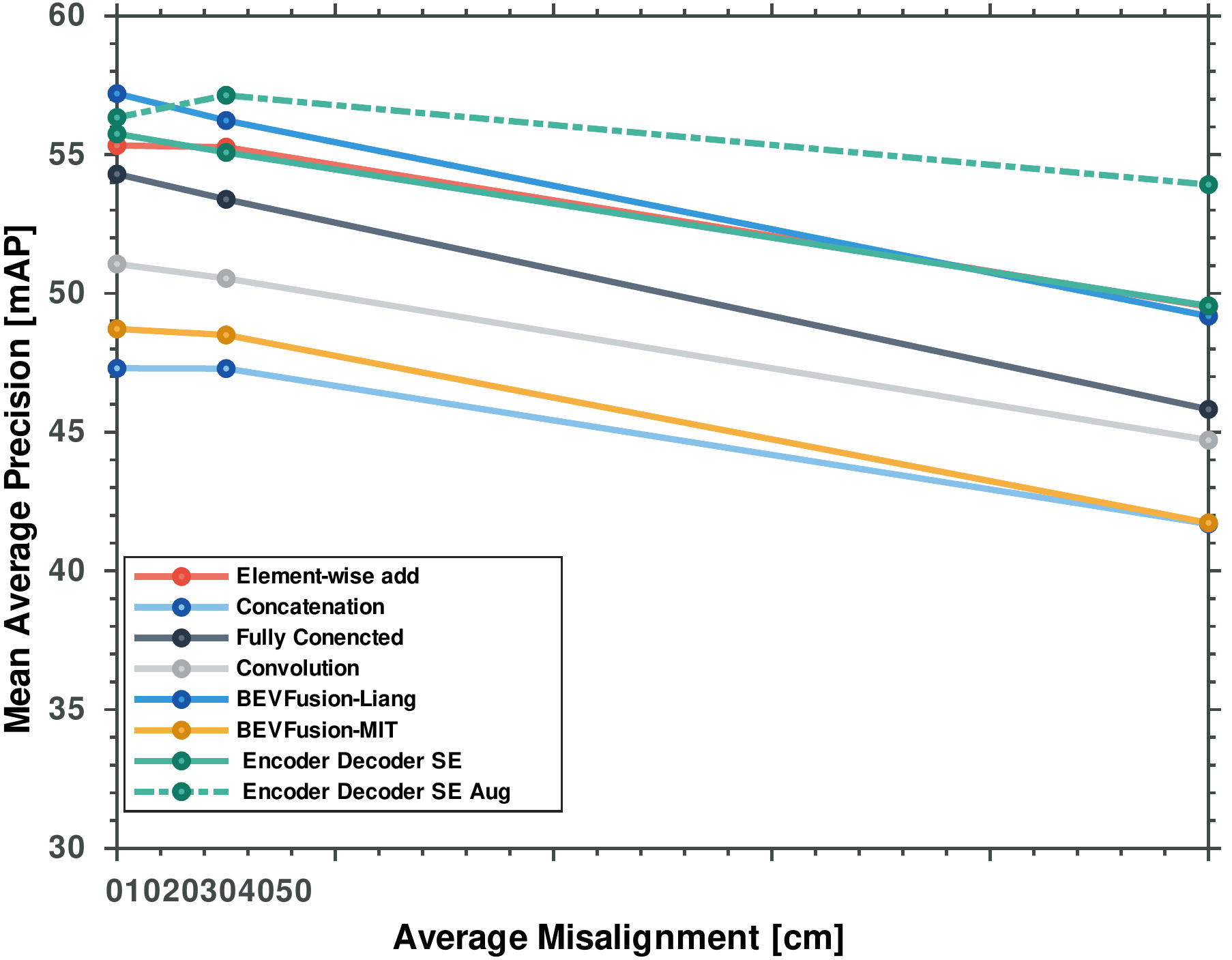}
         \caption{Sensor misalignment -- translation}
         \label{fig:viz_fusion_step_misalg_opt_trans}
     \end{subfigure}
     % \hfill
     \begin{subfigure}[b]{0.33\textwidth}
         \centering
         \includegraphics[width=\textwidth]{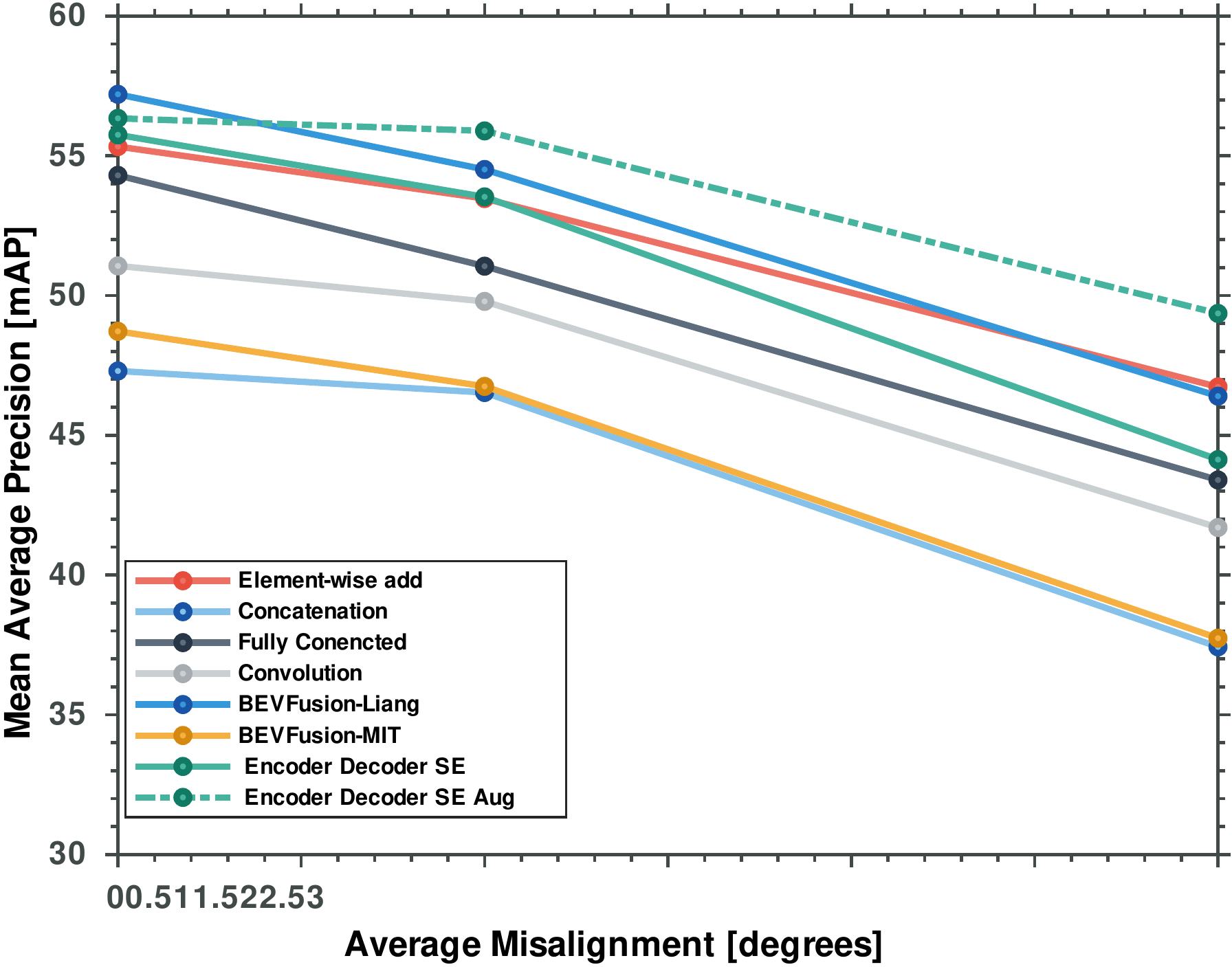}
         \caption{Sensor misalignment -- rotation}
         \label{fig:viz_fusion_step_misalg_opt_rots}
     \end{subfigure}
     \caption{Robustness evaluation of fusion steps on sensor misalignment.}
     \label{fig:benchmarkstep}
\end{figure*}

\subsection{Proposed fusion step - sensor misalignment}
\label{subsec:fusionstep_mis}

In real-world scenarios, sensors are often miscalibrated or decalibrate during the robot's movement. Fusion methods must account for these issues to be useful in real-world applications. The results of the sensor misalignment experiments are presented in Table \ref{tab:fusion_step_mis_alg_opt_results} and Figure \ref{fig:benchmarkstep}. During the training, the whole network (and fusion step) is trained for 6 epochs in addition to the LiDAR backbone which is pretrained for 24 epochs, and the image backbone is pretrained for 36 epochs. We trained and tested all the fusion steps on the \textit{NuScenes} data set. 
%and then tested them on the \textit{NuScenes Mini} subset. 

% , and the image stream is trained for 24 epochs
% The first learning strategy with a learning rate of $10^{-3}$. 
The training schedule uses an ADAM optimizer and involves step-downs in the learning rate at epoch 4 and epoch 6, the learning rate is lowered to $10^{-4}$, and then to $10^{-5}$ in the final epoch.

The convolution with encoder-decoder and SE-block marked with augmentation have the data abstraction applied during the training. This pipeline is \textit{identical} to the step-down learning schedule, except for the last epoch where a small amount of noise is added. The idea is that the noise makes the fusion step more general and less susceptible to noise.
% to the sensor transformations
The results of the misalignment experiments give insight into the performance of each fusion step in normal conditions, light, and finally severe misalignment.

% The first learning scheme resulted in largely worse performance over the entire set of models, thus we do not present them in this work. It is worth noting that the convolution fusion step achieved the highest performance with this training setup (max \textit{mAP} was $49.02$ and max \textit{NDS} $47.23$). 

The results shown in Table \ref{tab:fusion_step_mis_alg_opt_results}, highlight how the concatenation is at large the worst performer of the three simple fusion steps: convolution, element-wise add, and concatenation. We can also observe how the element-wise add is one of the best-performing fusion steps overall, despite its simple nature. Further, the convolution with SE-block is the best performer in the non-misaligned case, but our method, convolution with encoder-decoder and SE-block outperforms the other methods in small and large translational and rotational misalignments. 

Our augmented method performs better than the non-augmented version in all testing scenarios. Thus, we conclude that adding noise not only assists misaligned cases but also generalizes the fusion operation at large, achieving the best performance in almost every case on both corrupted and uncorrupted data. 

To summarize, our approach surpasses both the baseline and other state-of-the-art fusion step techniques when evaluated for sensor misalignment, exhibiting the lowest performance drop in both easy and hard cases.

% in the generalizing of

% as we can shown in \cref{subsec:fusionstep_mis},
% which can be explained by the shorter training process
%%%%%%%%%%%%%%%%%%%%%%%%%%%%%%%%%%%%%%%%%%%%%%%%%%%%%%%%%%%%%%%%%%%%%%%%%%%%%%%%%%%%%%%%%%%%%%%%%%%%%%%%%%%%%%%

\section{Discussion and future work}
% \textcolor{red}{add the paragraph about the distance and performance being lower because}

% that surpasses all of the existing fusion step methods for sensor misalignment.

% It performs better or on par with other methods when it comes to LiDAR layer reduction and point cloud downsampling. The key innovation of our fusion step lies in its robustness and adaptability to various scenarios real-world.

In this work, we have developed a novel fusion step for 3D object detection, robust and adaptable to various real-world scenarios. We compared our fusion step against a wide range of existing fusion methods. Our approach consistently outperformed these methods across different scenarios handling both corrupted and non-corrupted data. 

% \textcolor{red}{write about substituting it at different methods }

% One potential direction for future work is enhancing the fusion step's performance specifically in challenging weather conditions, such as rain or fog, that can significantly degrade the quality of both sensor inputs (LiDAR and camera), posing challenges for object detection algorithms. 

Although it demonstrates good performance, the improvement is limited due to the impact of sensor misalignment and lower resolution of the LiDAR on far away objects, compared to those nearby. In certain applications, such as mobile robots using lower-resolution LiDAR sensors, the solution could involve restricting LiDAR maximum distance during training and testing the methods since in these applications is more crucial to detect nearby objects than ones $50-70$m away. 

We could also consider assigning varying importance to features from the camera branch of the fusion pipeline based on the degree of LiDAR data corruption. However, it is important to note that this approach may not address the issue of sensor misalignment.

Finally, it would be beneficial to address challenging weather conditions, such as rain or fog, that can significantly degrade the quality of both sensor inputs. These questions will be deferred to future research, as they require further investigation and analysis.

\section{Acknowledgments}
This work was partially supported by the Wallenberg AI, Autonomous Systems and Software Program (WASP) funded by the Knut and Alice Wallenberg Foundation.

\appendix
% \label{subsec:model}

% Throughout the performance and robustness experiments showed in \cref{sec:experiments}, where the deep-fusion methods outperform the other methods in the LiDAR robustness experiments, we decided to use a 3D object detection model based on  BEVFusion-Liang~\cite{Bevfusion-Liang}, and perform further analysis and adaptation of different fusion steps using this model. 

We provide a short description of BEVFusion-Liang model, for more detail, please refer to the original paper~\cite{Bevfusion-Liang}.

The LiDAR branch of BEVFusion-Liang uses a SECOND~\cite{SECOND} as a Voxel Feature Encoding encoder, and a SECOND Feature Pyramid to create the PointPillars~\cite{PointPillars} backbone with the neck. In the image branch, a Composite Backbone Swin Transformer~\cite{SwinTransformer} is used, followed by the standard Feature Pyramid Network~\cite{lin2017feature} network neck.

 The features from the two backbones branches are represented in a tensor with dimensions:
\begin{equation}
    T_{LIDAR} = B \times W \times H \times C_1
\end{equation}
\begin{equation}
    T_{CAMERAS} = B \times W \times H \times C_2
\end{equation}

where, $B$ is the batch size, $W=200$ is the BEV feature space width, $H=200$ the BEV feature height, and, $C_1=384$, $C_2=256$ is the feature channels for the LiDAR feature set, and the RGB image set respectively. 

Following the fusion step is an anchor-based detection head~\cite{PointPillars}.

\bibliographystyle{ieeetr}
\bibliography{ref}
% \printbibliography
\end{document}